\documentclass{article}

\usepackage{microtype}
\usepackage{graphicx}
\usepackage{subcaption}
\usepackage{booktabs} %

\usepackage{hyperref}

\usepackage{booktabs}       %
\usepackage{multirow}
\usepackage{enumitem}       %
\usepackage{colortbl}       %
\definecolor{defaultcolor}{gray}{0.92}
\newcommand{\dft}{\rowcolor{defaultcolor}} %
\definecolor{linkblue}{RGB}{0,70,160} %

\newcommand{\boldstartspace}[1]{\vspace{0.2em}\noindent\textbf{#1}}

\usepackage{xspace}
\newcommand{\vs}{\textit{vs.}\xspace}
\newcommand{\eg}{\textit{e.g.}\xspace}
\newcommand{\ie}{\textit{i.e.}\xspace}

\usepackage[accepted]{icml2026}

\usepackage{amsmath}
\usepackage{amssymb}
\usepackage{mathtools}
\usepackage{amsthm}

\usepackage{graphicx}

\usepackage[capitalize,noabbrev]{cleveref}

\theoremstyle{plain}

\theoremstyle{definition}

\theoremstyle{remark}

\begin{document}

\twocolumn[
  \icmltitle{PointDiT: Pixel-Space Diffusion for Monocular Geometry Estimation}

  \begin{icmlauthorlist}
    \icmlauthor{Haofei Xu}{google,eth,tue}
    \icmlauthor{Rundi Wu}{google}
    \icmlauthor{Philipp Henzler}{google}
    \icmlauthor{Nikolai Kalischek}{google}
    \icmlauthor{Michael Oechsle}{google}
    \icmlauthor{Fabian Manhardt}{google}
    \icmlauthor{Marc Pollefeys}{eth,msft}
    \icmlauthor{Andreas Geiger}{tue,kesai}
    \icmlauthor{Federico Tombari}{google,tum}
    \icmlauthor{Michael Niemeyer}{google}
  \end{icmlauthorlist}

  \vskip 0.2in
  \begin{center}
  \href{https://haofeixu.github.io/pointdit}{\large \textcolor{linkblue}{https://haofeixu.github.io/pointdit}}
  \end{center}

  \icmlaffiliation{eth}{ETH Zurich}
  \icmlaffiliation{tue}{University of Tübingen, Tübingen AI Center}
  \icmlaffiliation{google}{Google}
  \icmlaffiliation{msft}{Microsoft}
  \icmlaffiliation{kesai}{KE:SAI}
  \icmlaffiliation{tum}{Technical University of Munich}

  \icmlcorrespondingauthor{Haofei Xu}{haofei.xu@inf.ethz.ch}

  \icmlkeywords{Machine Learning, ICML}

  \vskip 0.3in
]

\printAffiliationsAndNotice{}  %

\begin{abstract}
  State-of-the-art single-image 3D reconstruction methods often rely on complex hybrid architectures and loss functions, or compress geometry into latent spaces in order to leverage pre-trained latent diffusion models. In this work, we show that such architectural overhead and intricate loss formulations are unnecessary. We introduce a minimalist pixel-space Diffusion Transformer, built on a plain ViT, that operates directly on raw 3D point map patches and is conditioned on image tokens from a pre-trained DINOv3. Unlike existing latent diffusion approaches, we train our diffusion backbone entirely from scratch, eliminating the need for point map tokenizers. Despite its simplicity, our approach surpasses complex latent-based diffusion models while remaining significantly simpler than hybrid alternatives. Notably, it produces sharper geometric structure and is more robust in highly ambiguous regions, such as transparent objects.
\end{abstract}

\section{Introduction}

Monocular geometry estimation is a fundamental building block of 3D scene understanding, bridging 2D visual inputs and 3D spatial structure. In this work, we focus on predicting dense 3D point maps from single RGB images~\cite{wang2025moge,piccinelli2025unidepthv2}. Unlike depth maps, which capture only scalar distance and require camera intrinsics to recover 3D structure, point maps represent scene geometry directly in the camera coordinate system, enabling immediate 3D reconstruction without knowing the camera's calibration. However, mapping a single 2D image to a dense 3D representation remains ill-posed, owing to the inherent scale and depth ambiguities of perspective projection.

\begin{figure}[t]
    \centering
    \includegraphics[width=0.8\linewidth]{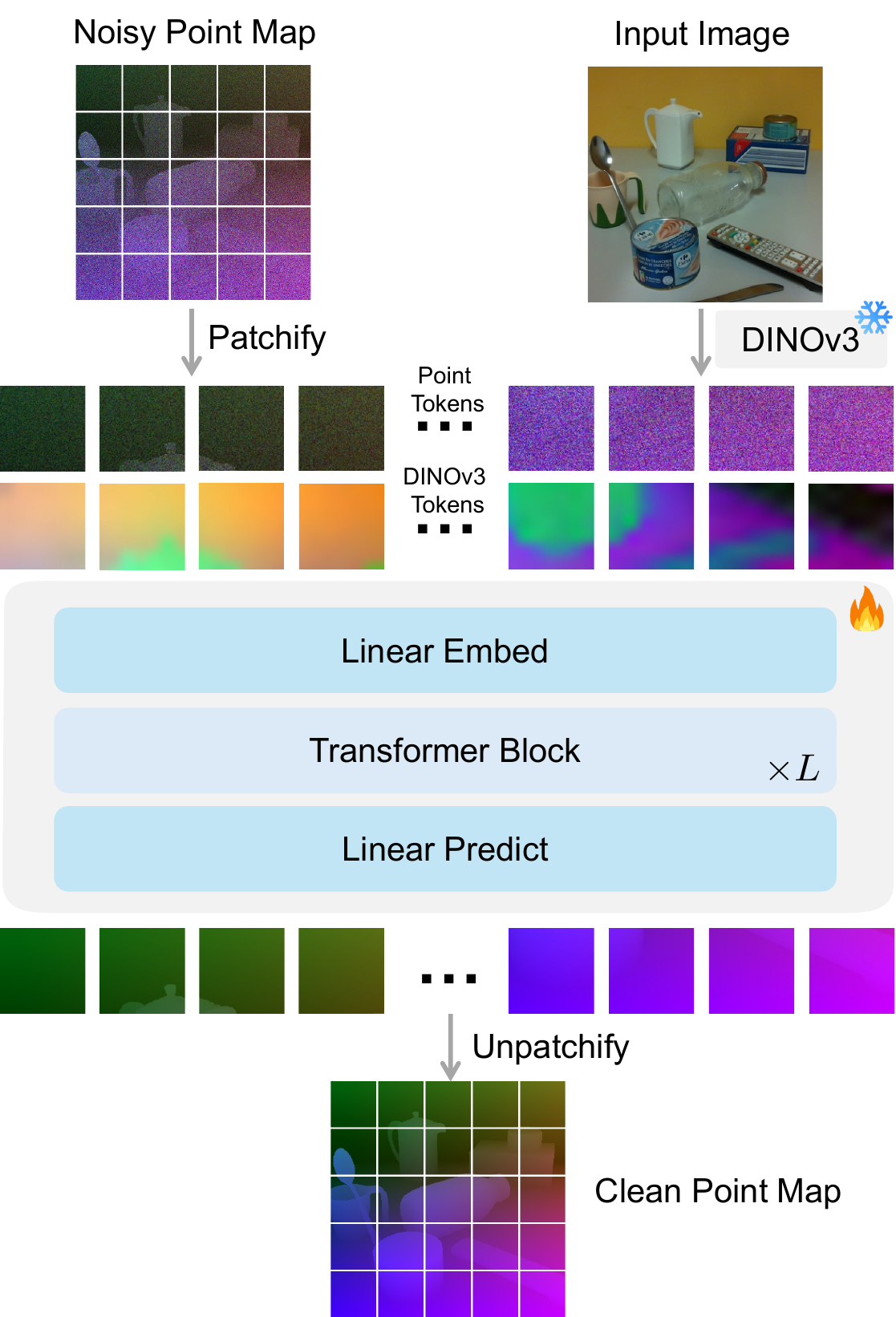}
    \caption{\textbf{PointDiT.} A minimalist pixel-space Diffusion Transformer operating directly on raw point map patches, conditioned on image tokens from a pre-trained DINOv3. The 3D point map ($H \times W \times 3$) is visualized as an RGB image, with color encoding the spatial $(X, Y, Z)$ coordinates.}
    \label{fig:teaser}
    \vspace{-2mm}
\end{figure}

\begin{figure}[t]
    \centering
    \begin{subfigure}{\linewidth}
        \centering
        \includegraphics[width=\linewidth]{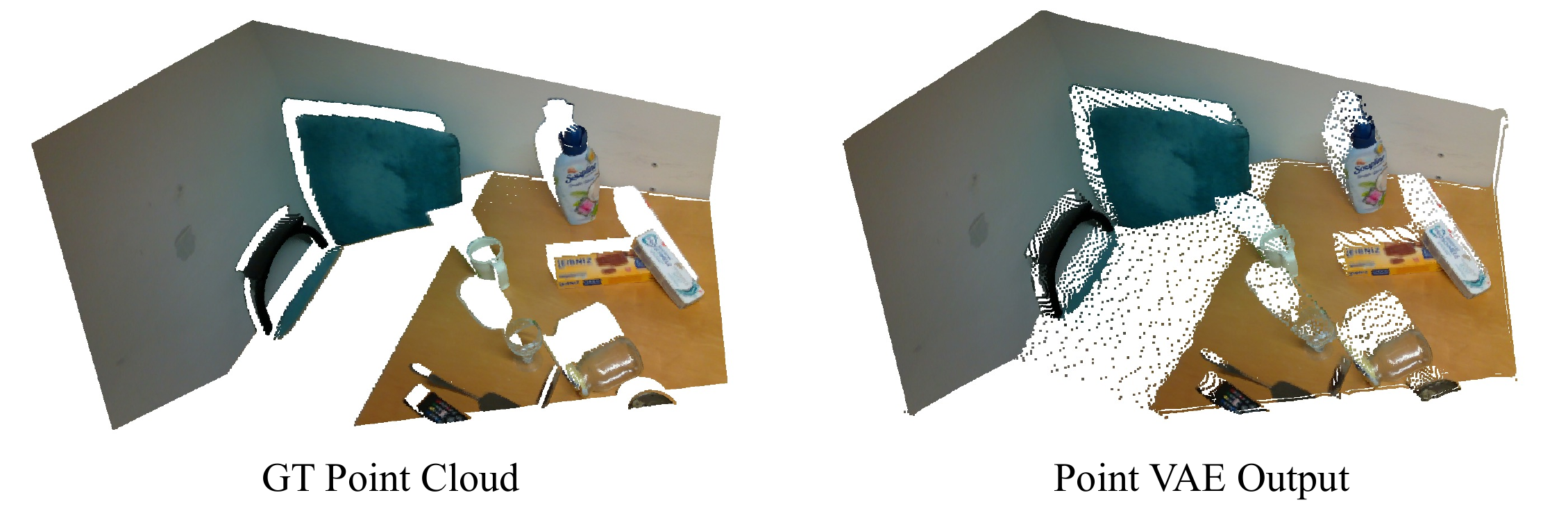}
        \caption{\textbf{Point VAE reconstruction.} Even without diffusion, the VAE-reconstructed point cloud exhibits substantial noise, which fundamentally limits the quality attainable by latent diffusion models.}
        \label{fig:vae_recon}
    \end{subfigure}

    \vspace{1mm}

    \begin{subfigure}{\linewidth}
        \centering
        \includegraphics[width=\linewidth]{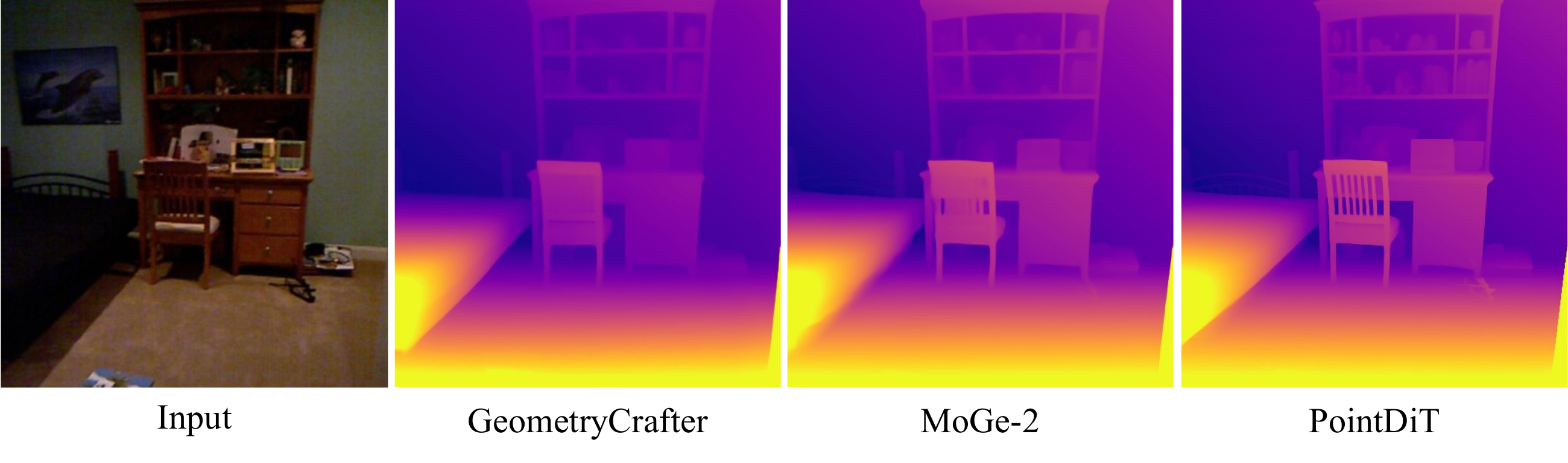}
        \caption{\textbf{Structural details.} PointDiT recovers intricate, thin structures such as the chair more faithfully than GeometryCrafter (latent diffusion) and MoGe-2 (deterministic regression). Here we visualize the $z$-depth from the predicted 3D point maps.}
        \label{fig:depth_details}
    \end{subfigure}

    \caption{\textbf{Comparison with latent diffusion and regression.} The two dominant paradigms each have an inherent limitation: (a) the VAE in latent diffusion models introduces reconstruction noise that caps the attainable quality, while (b) deterministic regression over-smooths fine geometric structures. PointDiT avoids both.}
    \label{fig:teaser_compare}
\end{figure}

Existing approaches to this challenge fall broadly into two categories. The first comprises deterministic regression models~\cite{yang2024depth,bochkovskii2025depth,piccinelli2025unidepthv2}. These methods often rely on complex hybrid architectures~\cite{wang2025moge,wang2025moge2,wang2025vggt,lin2025depth} that combine Vision Transformers (ViT)~\cite{dosovitskiy2020image} with convolutions~\cite{ranftl2021vision}, and require intricate loss functions~\cite{wang2025moge} to regularize training. Moreover, because of the task's inherent ambiguity, deterministic regressors tend to predict the mean of the output distribution, often yielding over-smoothed geometry that lacks high-frequency detail, particularly in complex scene regions (\cref{fig:depth_details}).

The second category seeks to resolve this ambiguity with Latent Diffusion Models (LDMs)~\cite{rombach2022high}, such as GeometryCrafter~\cite{xu2025geometrycrafter}. Although these methods exploit generative priors, they require compressing point maps into a latent space via a Variational Autoencoder (VAE)~\cite{kingma2013auto}. Building an expressive latent space for geometric data (\eg, point clouds) is non-trivial: their unbounded range and the relative scarcity of geometric data can limit the autoencoder's reconstruction quality and out-of-distribution generalization. In addition, constructing such a space often demands sophisticated tokenizer designs~\cite{xu2025geometrycrafter}. As a result, these methods frequently lose information during encoding and decoding, struggling to reconstruct fine geometric structures accurately (\cref{fig:vae_recon}). Furthermore, a fundamental trade-off exists between VAE reconstruction fidelity and diffusion generation capabilities due to their conflicting optimization objectives~\cite{bfl2025representation}, which inherently bounds the potential of latent diffusion models for geometric tasks.

In this work, we show that such architectural overhead and intricate loss formulations are unnecessary. Inspired by JiT~\cite{li2025back}, we introduce a minimalist pixel-space diffusion framework that trains directly on the raw point map space. This lets us exploit the probabilistic nature of diffusion to model ambiguous regions, without the signal degradation introduced by VAEs. Our architecture is simple by design: a plain Vision Transformer (ViT) operating on point map patches. A key element of our training recipe is the $x$-prediction objective, \ie, predicting the clean point map directly, rather than the $v$-prediction target commonly used in flow matching~\cite{salimans2022progressive,xu2025pixel}. Extending the findings of JiT~\cite{li2025back} beyond image generation, we show that this objective is highly effective for geometric data and yields substantially better point map estimation.

To guide geometry prediction, our diffusion model is conditioned on the input RGB image. Although it already works well with naive linear patchification~\cite{dosovitskiy2020image}, we find that injecting strong priors substantially improves performance. Specifically, we adopt DINOv3~\cite{simeoni2025dinov3} as a robust general-purpose feature extractor. Conditioning our plain ViT backbone on DINOv3 tokens bridges powerful priors from representation learning with diffusion training~\cite{yu2025representation,zheng2025diffusion}.

We show that this streamlined approach surpasses complex latent-based diffusion models~\cite{xu2025geometrycrafter} while remaining significantly simpler than hybrid deterministic alternatives~\cite{wang2025moge,lin2025depth}. Our model further excels at generating sharp geometric boundaries and resolving depth in highly ambiguous scenarios, such as transparent objects. PointDiT achieves highly competitive results with just one-step diffusion, and its structural details improve further with additional sampling steps. Beyond this specific task, our results suggest that pixel-space diffusion extends naturally beyond natural images to structured geometric signals such as 3D point maps, pointing toward a simpler paradigm for future 3D and 4D generation models.

\section{Related Work}

\paragraph{Latent Diffusion Models.}
Latent Diffusion Models (LDMs)~\cite{rombach2022high} have become the dominant paradigm for high-resolution image synthesis, decoupling the modeling of semantic content from perceptual detail by operating in a compressed latent space. Following this success, recent works adapt LDMs to geometric tasks~\cite{ke2024repurposing,he2024lotus,szymanowicz2025bolt3d,hu2025depthcrafter}. For instance, GeometryCrafter~\cite{xu2025geometrycrafter} and generative depth estimators~\cite{ke2024repurposing} use Variational Autoencoders (VAEs) to encode geometric maps into latent tokens. Although this reduces computational cost, the compression is fundamentally lossy: constructing a tokenizer that preserves the high-frequency precision required for 3D geometry is non-trivial, and standard image VAEs often smooth away fine structural details or introduce artifacts during decoding (\cref{fig:vae_recon}). In contrast, our approach bypasses the latent space entirely~\cite{li2025back}. By avoiding this architectural overhead and the associated information loss, we recover substantially sharper geometric boundaries (\cref{fig:depth_details}).

\paragraph{Pixel-Space Diffusion Models.}
Early diffusion models operate directly in pixel space~\cite{ho2020denoising}. Latent Diffusion Models (LDMs) subsequently shift generation into the latent space, and the Diffusion Transformer (DiT)~\cite{peebles2023scalable} replaces the conventional U-Net backbone with a Vision Transformer (ViT), achieving state-of-the-art class-conditional image generation. More recently, JiT~\cite{li2025back} shows that a ViT can be trained directly in pixel space, using direct data prediction ($x$-prediction) to cope with the high dimensionality of pixel space. While these advances have focused primarily on 2D image synthesis, we extend this minimalist pixel-space philosophy to dense 3D geometry estimation. By treating 3D point maps as multi-channel images and training a plain ViT backbone from scratch, we show that pixel-space diffusion is not only computationally feasible for 3D geometry but also superior to complex alternatives in reconstructing sharp details and resolving ambiguities.

\paragraph{Representation Learning and Generative Models.} A recent line of work connects representation learning with generative models. REPA~\cite{yu2025representation} and VA-VAE~\cite{yao2025reconstruction} observe that pre-trained vision encoders can dramatically improve generative diffusion models by regularizing their latent space. RAE~\cite{zheng2025diffusion} replaces the VAE in latent diffusion models with pre-trained representation autoencoders (\eg, DINOv2). At a high level, PointDiT shares this spirit, connecting DINOv3 with diffusion models. However, there are several key differences. First, RAE must be trained in two stages (reconstruction decoder and diffusion), whereas PointDiT is end-to-end. Second, RAE uses $v$-prediction, which requires scaling up the Transformer width, whereas PointDiT uses $x$-prediction, allowing us to train a smaller variant, PointDiT-B. Third, RAE requires 50 sampling steps for image synthesis, whereas PointDiT can perform one-step or few-step generation.

\paragraph{Monocular Depth Estimation.}
Estimating dense geometry from a single image is a longstanding problem in computer vision. Traditional discriminative approaches cast this as a regression task, using Convolutional Neural Networks (CNNs)~\cite{eigen2014depth} or, more recently, Transformers such as DPT~\cite{ranftl2021vision} and Depth Anything~\cite{yang2024depth} to predict scalar depth maps. However, depth maps are 2.5D representations that require known camera intrinsics to be lifted into 3D, which are often unavailable in unconstrained settings. Generative approaches such as Marigold~\cite{ke2024repurposing} instead repurpose pre-trained image diffusion models (\eg, Stable Diffusion) for depth estimation. Although these methods exploit strong generative priors, they remain fundamentally limited by the quality of VAEs. Moreover, there is often a trade-off between VAE reconstruction and diffusion generation, and balancing the two requires additional effort~\cite{bfl2025representation}. More recently, PPD~\cite{xu2025pixel} applies pixel-space diffusion to monocular depth estimation. However, PPD still uses the $v$-prediction target, which performs worse than $x$-prediction in our controlled comparisons (\cref{tab:ablation}(a)).

\paragraph{Monocular Point Map Estimation.}
To overcome the limitations of scalar depth, point map estimation predicts dense 3D coordinates $xyz$ directly in the camera coordinate system, enabling holistic 3D reconstruction without intrinsic calibration. The current state of the art is dominated by deterministic feed-forward models such as MoGe~\cite{wang2025moge,wang2025moge2}. These methods typically employ complex hybrid architectures that fuse ViTs with convolutional layers and rely on intricate loss functions to enforce geometric consistency. Despite their efficacy, deterministic regressors suffer from the inherent ambiguity of monocular projection, tending to predict the mean of the distribution. This often yields over-smoothed geometry, particularly in regions of high uncertainty or transparency (\cref{fig:vs_regression_viz}). We address this by casting point map estimation as a probabilistic generation task, allowing our model to capture sharp, high-frequency detail that deterministic baselines fail to resolve.

\section{Approach}
\label{sec:method}

We address dense point map prediction from a single RGB image. Formally, given an input image $\mathbf{c} \in \mathbb{R}^{H \times W \times 3}$, our goal is to estimate the corresponding point map $\mathbf{x} \in \mathbb{R}^{H \times W \times 3}$, in which each pixel encodes its 3D spatial ($X$, $Y$, $Z$) coordinates. To model the inherent ambiguities of this single-image setting, we propose a flow matching framework parameterized by a Vision Transformer (ViT)~\cite{dosovitskiy2020image,peebles2023scalable}. Our method learns to transport a simple Gaussian noise distribution to the data distribution of point maps, conditioned on the input image.

\subsection{Point Map Generation with Flow Matching}
\paragraph{Flow Matching.} We adopt the flow matching formulation to model point map generation from a single image. Flow matching learns an Ordinary Differential Equation (ODE) that continuously transforms a prior noise distribution $p_0$ into the data distribution $p_1$.

Let $\mathbf{z}_t$ denote the state at time $t \in [0, 1]$, defined by a linear interpolation between a noise sample $\boldsymbol{\epsilon} \sim p_0 = \mathcal{N}(\mathbf{0}, \mathbf{I})$ and a ground-truth data sample $\mathbf{x} \sim p_1$:
\begin{equation}
    \mathbf{z}_t = t \cdot \mathbf{x} + (1 - t) \cdot \boldsymbol{\epsilon}.
    \label{eq:interpolation}
\end{equation}
In this formulation, $t=0$ corresponds to pure noise ($\mathbf{z}_0 = \boldsymbol{\epsilon}$) and $t=1$ to the clean data ($\mathbf{z}_1 = \mathbf{x}$). The vector field $\mathbf{v}_t(\mathbf{z}_t)$ that generates this probability path is given by the time derivative of the state:
\begin{equation}
    \mathbf{v}_t = \frac{d\mathbf{z}_t}{dt} = \mathbf{x} - \boldsymbol{\epsilon}.
    \label{eq:velocity_def}
\end{equation}
This linear path induces a constant velocity for each sample pair $(\mathbf{x}, \boldsymbol{\epsilon})$, ensuring straight-line transport between the noise and data distributions.

\paragraph{Image Conditioned Flow Matching.} We extend this framework to model the conditional distribution $p(\mathbf{x} | \mathbf{c})$, where $\mathbf{c}$ is the input RGB image and $\mathbf{x}$ is the target dense point map. Specifically, we learn a conditional vector field $\mathbf{v}_\theta(\mathbf{z}_t, t | \mathbf{c})$ that predicts the target velocity defined in~\cref{eq:velocity_def}. By conditioning on $\mathbf{c}$, the model exploits the image's spatial context to resolve geometric ambiguity, steering the flow toward the target point map.

\paragraph{Point Map Normalization.}
Unlike standard RGB images bounded within $[0, 1]$, dense point maps exhibit varying coordinate ranges depending on the scene domain (\eg, indoor \vs outdoor scenes). In our flow matching formulation, the training target relies on the interpolation $\mathbf{z}_t = t \mathbf{x} + (1-t) \boldsymbol{\epsilon}$, where the noise $\boldsymbol{\epsilon}$ follows a fixed standard normal distribution $\mathcal{N}(\mathbf{0}, \mathbf{I})$. If the scale of the point data $\mathbf{x}$ far exceeds that of the noise, the data signal dominates the interpolation path even at near-zero time steps. This prevents the noise from destroying the data structure, destabilizing diffusion training. To mitigate this, we standardize the point maps before training. For each point map, we compute the centroid $\boldsymbol{\mu}$ and a scalar scale factor $s$, defined as the mean Euclidean distance of the points from the centroid. The normalized data $\tilde{\mathbf{x}}$ is given by:
\begin{equation}
    \tilde{\mathbf{x}} = \frac{\mathbf{x} - \boldsymbol{\mu}}{s}.
\end{equation}
This normalization brings the data to a scale comparable to the noise prior, facilitating stable flow matching. Our model is trained in this normalized space, and thus its point map predictions are affine-invariant, \ie, recovered up to an unknown scale and shift.

\paragraph{Sky Processing.}\label{par:sky} To accommodate the effectively infinite depth of the sky in outdoor scenes, we exclude sky points when computing the normalization statistics $\boldsymbol{\mu}$ and $s$, and then, in the resulting normalized frame, project them onto a virtual sphere of fixed radius $3$ (corresponding to $3\sigma$ of the standard normal noise prior). Since this radius is only a synthetic proxy for the true depth, we down-weight sky pixels in the training loss rather than masking them out entirely (\cref{sec:training}). Retaining a small but nonzero weight keeps the sky supervised as a distant background, so its geometry does not become arbitrary in the absence of supervision, while preventing its pseudo-depth values from dominating the optimization. This, in turn, enables stable joint training across heterogeneous indoor and outdoor datasets. At inference, we discard predicted 3D points whose norm exceeds $2.9$, a small margin below the sky sphere of radius $3$.

\subsection{Architecture}
We implement $F_\theta$ as a Vision Transformer (ViT), which serves as our pixel-space Diffusion Transformer. The network takes the noisy point map $\mathbf{z}_t$, the current time step $t$, and the conditioning image $\mathbf{c}$ as input. Crucially, unlike previous flow matching models that typically predict the velocity, our network is trained to predict the clean point map. Inspired by JiT~\cite{li2025back} for 2D image generation, we show that this clean data prediction target is likewise crucial for 3D point map data. \cref{fig:teaser} shows an overview of our architecture.

\paragraph{Point Map Patchification.}
The noisy point map $\mathbf{z}_t \in \mathbb{R}^{H \times W \times 3}$ has the same spatial resolution as the input image, so applying a ViT directly at the pixel level would be prohibitively expensive. We therefore patchify it, partitioning the map into a regular grid of non-overlapping $p \times p$ patches to reduce the ViT sequence length from pixels to patches. This yields $N = (H/p) \times (W/p)$ patches, each flattened into a vector of size $3p^2$. These vectors are then mapped to the embedding dimension $D$ by a learnable linear projection $\phi$, giving the point map tokens $\mathbf{T}_z = \phi(\text{Patchify}(\mathbf{z}_t)) \in \mathbb{R}^{N \times D}$.

\paragraph{Image Conditioning.}
The conditioning image $\mathbf{c}$ provides structural guidance for generation. Since $\mathbf{c}$ is clean, unlike the noisy $\mathbf{z}_t$, we can exploit powerful pre-trained representations to encode it. Although a standard learnable linear patch embedding~\cite{dosovitskiy2020image} already works well, we find that extracting patch tokens with a frozen DINOv3 encoder~\cite{simeoni2025dinov3} leads to better performance. Unlike RAE~\cite{zheng2025diffusion}, which uses only the last layer, we find it beneficial to combine DINOv3 features from multiple layers. In particular, we use four uniformly spaced intermediate layers, following the layer selection of the DPT~\cite{ranftl2021vision} head. Unlike DPT, which relies on sophisticated convolutions to fuse these features, we simply concatenate the tokens along the channel dimension, capturing a rich feature hierarchy that ranges from low-level details to high-level abstractions. This yields a composite image representation $\mathbf{T}_c \in \mathbb{R}^{N \times 4D}$, where $D$ is the per-layer feature dimension. To ensure spatial alignment, we use the same patch size $p=16$ and embedding dimension $D$ for both the point map and DINOv3 branches.

\paragraph{Image and Point Map Fusion.}
Given the spatial alignment between $\mathbf{T}_c$ and $\mathbf{T}_z$, we fuse the two modalities by channel-wise concatenation, forming the input $\mathbf{T}_{\text{in}} = \text{Concat}(\mathbf{T}_c, \mathbf{T}_z) \in \mathbb{R}^{N \times 5D}$ to the Diffusion Transformer. A linear layer projects this from $5D$ to the embedding dimension $D$. The sequence is then processed by a stack of Transformer blocks~\cite{dosovitskiy2020image,li2025back}, each comprising multi-head self-attention and an MLP.

\paragraph{Clean Point Map Prediction.}
The Diffusion Transformer outputs a sequence of refined tokens $\mathbf{T}_{\text{out}} \in \mathbb{R}^{N \times D}$. To recover the dense point map, we apply a linear prediction head that projects each token from $D$ back to the flattened patch dimension $3p^2$, yielding patch vectors in $\mathbb{R}^{N \times 3p^2}$. An unpatchification operation then rearranges these vectors into the original spatial grid $(H/p) \times (W/p) \times p \times p \times 3$ and permutes the dimensions to form the full-resolution tensor $\hat{\mathbf{x}} \in \mathbb{R}^{H \times W \times 3}$. This tensor is the model's estimate of the clean point map at the current step.

\subsection{Training}
\label{sec:training}

\paragraph{Noise Schedule.}
To sample the time step $t \in [0, 1]$ during training, we use a logit-normal distribution~\cite{esser2024scaling}, following JiT~\cite{li2025back}. Specifically, we draw $z \sim \mathcal{N}(\mu, \sigma^2)$ with $\mu = -0.8$ and $\sigma = 0.8$, and map it to the time domain through the sigmoid function, $t = \text{sigmoid}(z)$.

In our point map generation task, we observe that the sigmoid only asymptotically approaches its boundaries, so the model is never trained on the exact pure-noise state ($t=0$).
This creates a train-test discrepancy, since inference always starts at $t=0$, and the model may then struggle to initiate the flow trajectory from the prior~\cite{lin2024common}. To resolve this, we adopt a rectified sampling strategy: with probability $p_{\text{zero}} = 0.1$, we override the logit-normal sample and set $t=0$ explicitly. This calibrates the model to the pure-noise distribution it encounters at the start of inference.

\paragraph{Velocity Loss.}
Although our network $F_\theta$ is parameterized to predict the clean point map $\hat{\mathbf{x}}$, we optimize it in velocity space ($v$-loss), following JiT~\cite{li2025back}. In our experiments, this performs slightly better than computing the loss directly on $\hat{\mathbf{x}}$ ($x$-loss). During training, we construct the noisy input $\mathbf{z}_t$ (\cref{eq:interpolation}) and obtain the network prediction $\hat{\mathbf{x}} = F_\theta(\mathbf{z}_t, t, \mathbf{c})$. We then convert $\hat{\mathbf{x}}$ into an estimated velocity $\hat{\mathbf{v}}_t$ by rearranging the interpolation path:
\begin{equation}
    \hat{\mathbf{v}}_t(\mathbf{z}_t, t) = \frac{\hat{\mathbf{x}} - \mathbf{z}_t}{1 - t}.
    \label{eq:v_prediction}
\end{equation}
To ensure numerical stability as $t \to 1$, we clip the denominator $(1-t)$ to a minimum threshold $\delta = 0.05$.

We minimize the Mean Squared Error (MSE) between this estimated velocity $\hat{\mathbf{v}}_t$ and the constant ground-truth velocity target $\mathbf{u}_t = \mathbf{x} - \boldsymbol{\epsilon}$:
\begin{equation}
    \mathcal{L}_{\text{fm}} = \mathbb{E}_{\mathbf{x}, t, \boldsymbol{\epsilon}} \left[ \frac{1}{M} \sum_{i=1}^{M} w_i \left\| \hat{\mathbf{v}}_{t,i} - (\mathbf{x}_i - \boldsymbol{\epsilon}_i) \right\|_2^2 \right],
    \label{eq:fm_loss}
\end{equation}
where $i$ indexes the $M$ pixels and $w_i$ is a per-pixel weight that down-weights sky pixels ($w_i = 0.01$ for sky pixels and $w_i = 1$ otherwise), as motivated in \cref{par:sky}.

\paragraph{Relative Point Loss.}
The flow matching loss alone already yields an effective model. Nonetheless, because our model predicts the clean output directly in the original data space, without a VAE, it is straightforward to impose auxiliary losses on this output when necessary. To show this flexibility, we add a relative point loss. Point maps span a high dynamic range: distant points have large coordinate norms that dominate standard error metrics, leaving nearby details under-weighted. We therefore normalize each per-pixel error by the magnitude of the target point, which emphasizes the reconstruction of local detail:
\begin{equation}
\label{eq:rel_point}
    \mathcal{L}_{\text{rel}} = \mathbb{E}_{\mathbf{x}, t, \boldsymbol{\epsilon}} \left[\frac{1}{M} \sum_{i=1}^{M} w_i \frac{\| \hat{\mathbf{x}}_i - \mathbf{x}_i \|_1}{\| \mathbf{x}_i \|_2 + \xi}\right],
\end{equation}
where $w_i$ is the same per-pixel sky weight as in \cref{eq:fm_loss} and $\xi$ is a small stability constant.

\paragraph{Total Loss.}
The final optimization objective is the weighted sum:
\begin{equation}
    \mathcal{L} = \mathcal{L}_{\text{fm}} + \lambda \mathcal{L}_{\text{rel}},
\end{equation}
where $\lambda=0.1$ is the loss weight. We train the full model end-to-end. Unlike existing methods (\eg, MoGe~\cite{wang2025moge}) that typically rely on several regularization losses, our training is driven primarily by the flow matching loss, with only a single lightweight auxiliary term.

\subsection{Inference}
During inference, we recover $\mathbf{x}$ from pure noise $\mathbf{z}_0 \sim \mathcal{N}(\mathbf{0}, \mathbf{I})$, conditioned on the input image $\mathbf{c}$, by solving the ODE $d\mathbf{z}_t = \mathbf{v}_\theta(\mathbf{z}_t, t \mid \mathbf{c})\, dt$ from $t=0$ to $t=1$. We use a standard Euler solver with step size $\Delta t$. At each step $t$, we predict the clean data $\hat{\mathbf{x}}$, derive the velocity $\hat{\mathbf{v}}_t$, and update the state:
\begin{equation}
    \mathbf{z}_{t + \Delta t} \leftarrow \mathbf{z}_t + \Delta t \cdot \hat{\mathbf{v}}_t.
\end{equation}
This iterative process transports the sample along the learned linear trajectory to reconstruct the final point map.

Surprisingly, our model can perform single-step inference with competitive results, while additional steps further improve quality using the same model (\cref{fig:details_vs_steps}). We attribute this to the per-pixel alignment between the predicted point map and the conditioning image: each output location is largely determined by its corresponding image feature, making the transport from noise to the target geometry nearly a direct mapping that is accurate even in one step, with additional steps mainly refining details. We further observe that our model can serve as a deterministic estimator at inference time, by initializing from all zeros instead of random noise (\cref{tab:infer_noise_sample}). A similar behavior was reported for diffusion-based depth estimation~\cite{garcia2025fine}, likely because the model learns to be robust to the input noise, or even to constant zeros.

\section{Experiments}

\subsection{Datasets}

We adopt a two-stage training strategy for efficient training. The model is first pre-trained at $256 \times 256$ resolution and then fine-tuned at $512 \times 512$, using only synthetic data throughout. For $256 \times 256$ pre-training, we use SceneNet-RGBD~\cite{mccormac2017scenenet}, which provides approximately 5.36 million photorealistic RGB-D samples. The $512 \times 512$ fine-tuning stage uses a high-fidelity mixture of 11 synthetic datasets: Hypersim~\cite{roberts2021hypersim}, VKITTI2~\cite{cabon2020virtual}, UrbanSyn~\cite{gomez2025all}, Synscapes~\cite{wrenninge2018synscapes}, TartanAir~\cite{wang2020tartanair}, OmniWorldGame~\cite{zhou2025omniworld}, EDEN~\cite{le2021eden}, IRS~\cite{wang2019irs}, Dynamic Replica~\cite{karaev2023dynamicstereo}, MVSSynth~\cite{huang2018deepmvs}, and TartanGround~\cite{wang2025tartanground}, totaling approximately 6.22 million samples. As all of these datasets are RGB-D, we convert their raw depth maps into point maps using the provided camera intrinsics. More dataset details are provided in the appendix (\cref{tab:datasets}).

We train exclusively on synthetic data for two reasons. 1) Geometric precision: synthetic environments provide ``pixel-perfect'' ground-truth point maps, which are essential for learning high-quality, dense 3D distributions. 2) Domain agnosticism: because our architecture models the underlying geometry (point maps) rather than low-level image textures, the synthetic-to-real appearance gap matters less for our task. To further bridge the gap between synthetic and real-world distributions, we incorporate frozen features from a pre-trained DINOv3 backbone. These self-supervised representations provide robust, domain-invariant visual cues that allow our model to focus on geometric reconstruction while generalizing to natural images.

\subsection{Implementation Details}

\boldstartspace{Model Configurations.} We implement three scale variants of our architecture: PointDiT-B (Base), PointDiT-L (Large), and PointDiT-H (Huge), following the configurations of JiT~\cite{li2025back}. For the visual backbone, we use a frozen DINOv3 encoder to extract patch-level embeddings, scaling its capacity with each variant (\eg, ViT-L features for PointDiT-L). Apart from this frozen encoder, all Transformer layers and prediction heads are trained from scratch. We use the same patch size of 16 for all variants.

\boldstartspace{Training Schedule.}
Our training curriculum consists of large-scale pre-training followed by high-resolution fine-tuning. We use the AdamW optimizer~\cite{loshchilov2017decoupled}, with a learning rate schedule and hyperparameters consistent with JiT~\cite{li2025back}. More implementation details are provided in the appendix (\cref{sec:train_details}). 

All variants are pre-trained at $256 \times 256$ for 30 epochs (including a 5-epoch warmup) and then fine-tuned at $512 \times 512$, scaling the number of GPUs with resolution and model capacity. Interestingly, we observe that larger models converge faster and require fewer fine-tuning epochs. We report the detailed per-variant training cost (GPU count and wall-clock time) in the appendix (\cref{tab:training}).

\begin{table}[t]

\begin{center}
\caption{\textbf{Comparisons}. Average results on 7 real-world evaluation datasets with 3,444 samples. The image resolution is $512 \times 512$. Rel$^\text{p}$ and $\delta_1^\text{p}$ are point map metrics, while Rel$^\text{d}$ and $\delta_1^\text{d}$ are depth map metrics. BF1 measures boundary sharpness. PointDiT-H attains the best depth accuracy (Rel$^\text{d}$ and $\delta_1^\text{d}$) and PointDiT the sharpest boundaries (BF1) among all methods, while being far more efficient than the latent diffusion model GeometryCrafter ($72$ \vs $1{,}178$~ms for single-step inference). Even a single sampling step already surpasses all prior methods on BF1, and additional steps further sharpen boundaries at a modest cost.
    }
    \setlength{\tabcolsep}{2pt} %
    \resizebox{\linewidth}{!}{
    \begin{tabular}{lcccccccccccccccccccccccc}
    \toprule
    Method & Rel$^\text{p}$ $\downarrow$ & $\delta_1^\text{p}$ $\uparrow$ & Rel$^\text{d}$ $\downarrow$ & $\delta_1^\text{d}$ $\uparrow$ & BF1 $\uparrow$ & {\begin{tabular}[x]{@{}c@{}}Param\\ (M) \end{tabular}} & {\begin{tabular}[x]{@{}c@{}}Time\\ (ms) \end{tabular}} \\

    \midrule
    GeometryCrafter & 5.45 & 96.75 & 3.52 & 97.84 & 4.64 & 1,937 & 1,178 \\
    PPD & 5.54 & 96.59 & 3.88 & 97.78 & 9.28 & 804 & 402 \\
    Depth Pro & 5.71 & 96.71 & 3.84 & 97.63 & 9.41 & 952 & 68 & \\
    UniDepthV2 & 4.45 & 97.35 & 2.86 & 98.52 & 6.94 & 354 & 26 \\
    DA3 & 4.77 & 96.63 & 3.22 & 97.81 & 6.33 & 1,356 & 82 \\
    MoGe & \textbf{4.21} & 97.45 & 3.10 & 98.01 & 5.61 & 314 & 34 \\
    MoGe-2 & 4.53 & 97.46 & 2.90 & 98.45 & 7.40 & 326 & 24 \\
    
    \midrule
    PointDiT-B (1 step) & 5.84 & 96.71 & 3.70 & 97.84 & 8.18 & 223 & 31 \\
    PointDiT-B (2 steps) &	5.81	& 96.77	& 3.64	& 97.86	& 8.88	& 223	& 47 \\
    PointDiT-B (3 steps) &	5.83 & 96.79 & 3.64 & 97.86 & 9.09 & 223 & 63 \\
    PointDiT-B (4 steps) &	5.85	& 96.80	& 3.64	& 97.86	& 9.16	& 223	& 79 \\
    \midrule
    PointDiT-L (1 step) & 4.90 & 97.42 & 3.15 & 98.22 & 9.56 & 771 & 65 \\
    PointDiT-L (2 steps) &	4.84	& 97.52	& 3.09	& 98.24	& 10.11	& 771	& 87 \\
    PointDiT-L (3 steps) &	 4.85 & 97.54 & 3.09 & 98.25 & 10.36 & 771 & 109 \\
    PointDiT-L (4 steps)	& 4.85	& 97.55	& 3.09	& 98.25	& \textbf{10.50}	& 771	& 131 \\
    \midrule
    PointDiT-H (1 step) & 4.45 & 97.93 & 2.81 & 98.51 & 9.79 & 1,807 &72 \\
    PointDiT-H (2 steps) & 4.38 & 97.99 & \textbf{2.75} & \textbf{98.54} & 10.31 & 1,807 &116 \\
    PointDiT-H (3 steps) & 4.39 & 98.01 & \textbf{2.75} & \textbf{98.54} & 10.44 & 1,807 &160 \\
    PointDiT-H (4 steps) & 4.40 & \textbf{98.02} & \textbf{2.75} & \textbf{98.54} & 10.49 & 1,807 &204 \\
    
    \bottomrule
    \end{tabular}
    }

    \label{tab:comparison}
    \end{center}
    \vspace{-6mm}
\end{table}

\subsection{Evaluation Setup and Metrics}

To assess the zero-shot generalization of our model, we evaluate on seven commonly used real-world datasets: DIODE~\cite{vasiljevic2019diode}, KITTI~\cite{geiger2012kitti}, NYUv2~\cite{silberman2012nyuv2}, ETH3D~\cite{schops2017eth3d}, HAMMER~\cite{jung2022hammer}, iBims-1~\cite{koch2018ibims}, and Booster~\cite{zamaramirez2022booster}. These span diverse environments, from indoor rooms to complex outdoor driving scenes. More details are provided in the appendix (\cref{tab:eval_datasets}). Consistent with our training, we evaluate at both $256 \times 256$ and $512 \times 512$ resolutions. Given the heterogeneous aspect ratios and resolutions of the original test sets, we adopt a standardized preprocessing pipeline: each input image is rescaled so that its shorter side matches the target resolution (256 or 512 pixels) and then center-cropped to the square input required by the model. For a fair comparison, we benchmark against several state-of-the-art baselines, evaluating their publicly available pre-trained weights under the same preprocessing and cropping protocol.

Our model predicts affine-invariant point maps, from which affine-invariant depth maps are obtained by extracting the $z$-component of each point. For evaluation, we follow the alignment procedure of MoGe~\cite{wang2025moge}, determining the optimal scale and shift by solving a least-squares problem that minimizes the discrepancy between the prediction and the ground truth. We assess prediction quality in both the point map and depth domains using standard metrics~\cite{wang2025moge}:

\begin{itemize}[itemsep=2pt, topsep=2pt, parsep=0pt, leftmargin=*]
    \item \textbf{Accuracy} ($\delta_1$): the percentage of pixels for which the ratio between prediction and ground truth is below $1.25$.
    \item \textbf{Relative absolute error} (Rel): $\frac{1}{N} \sum \frac{|y - \hat{y}|}{y}$, the scale-normalized error.
    \item \textbf{Geometric edge fidelity} (BF1): a local boundary metric, following Depth Pro~\cite{bochkovskii2025depth}, that assesses the recovery of fine structural details and sharp depth discontinuities.
\end{itemize}

We report metrics in both domains, using Rel$^\text{p}$ and $\delta_1^\text{p}$ for point maps and Rel$^\text{d}$ and $\delta_1^\text{d}$ for depth.

\begin{table}[t]

\begin{center}
\caption{\textbf{Single-step feed-forward inference}. Single-step results of PointDiT-H from random noise (three seeds) or an all-zeros input. Performance is nearly invariant to the noise, with all-zeros matching or slightly exceeding stochastic sampling, indicating the model learns to be robust to different noise realizations.
    }
    \setlength{\tabcolsep}{2pt} %
    \begin{tabular}{lcccccccccccccccccccccccc}
    \toprule
    Method & Rel$^\text{p}$ $\downarrow$ & $\delta_1^\text{p}$ $\uparrow$ & Rel$^\text{d}$ $\downarrow$ & $\delta_1^\text{d}$ $\uparrow$ & BF1 $\uparrow$ \\

    \midrule

    rand noise (seed 1) & 4.454 & 97.928 & 2.815 & 98.505 & 9.772 \\
    rand noise (seed 2) & 4.452 & 97.938 & 2.811 & 98.513 & 9.778 \\
    rand noise (seed 3) & 4.454 & 97.921 & 2.812 & 98.513 & 9.772 \\
    all zeros (no rand) & 4.446 & 97.934 & 2.806 & 98.508 & 9.792 \\

    \bottomrule
    \end{tabular}

    \label{tab:infer_noise_sample}
    \end{center}
    \vspace{-6mm}
\end{table}

\begin{figure}[t]
    \centering
    \includegraphics[width=\linewidth]{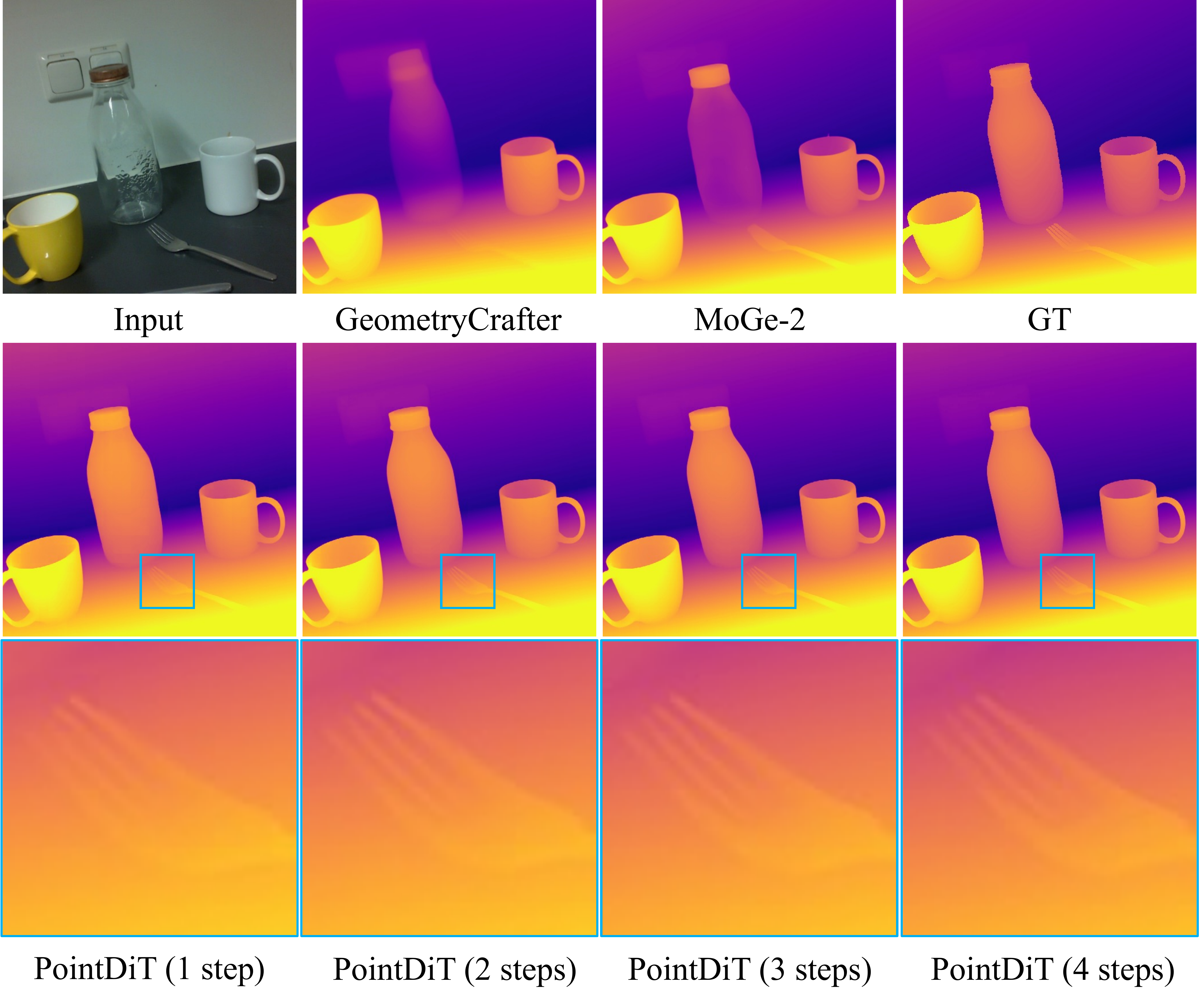}
    \caption{\textbf{Different diffusion sampling steps.} Our single-step diffusion already significantly outperforms prior works, and increasing the sampling steps further enhances reconstruction details (see the zoomed-in region). } 
    \label{fig:details_vs_steps}
    \vspace{-2mm}
\end{figure}

\begin{figure*}[t]
    \centering
    \vspace{-2mm}
    \includegraphics[width=\linewidth]{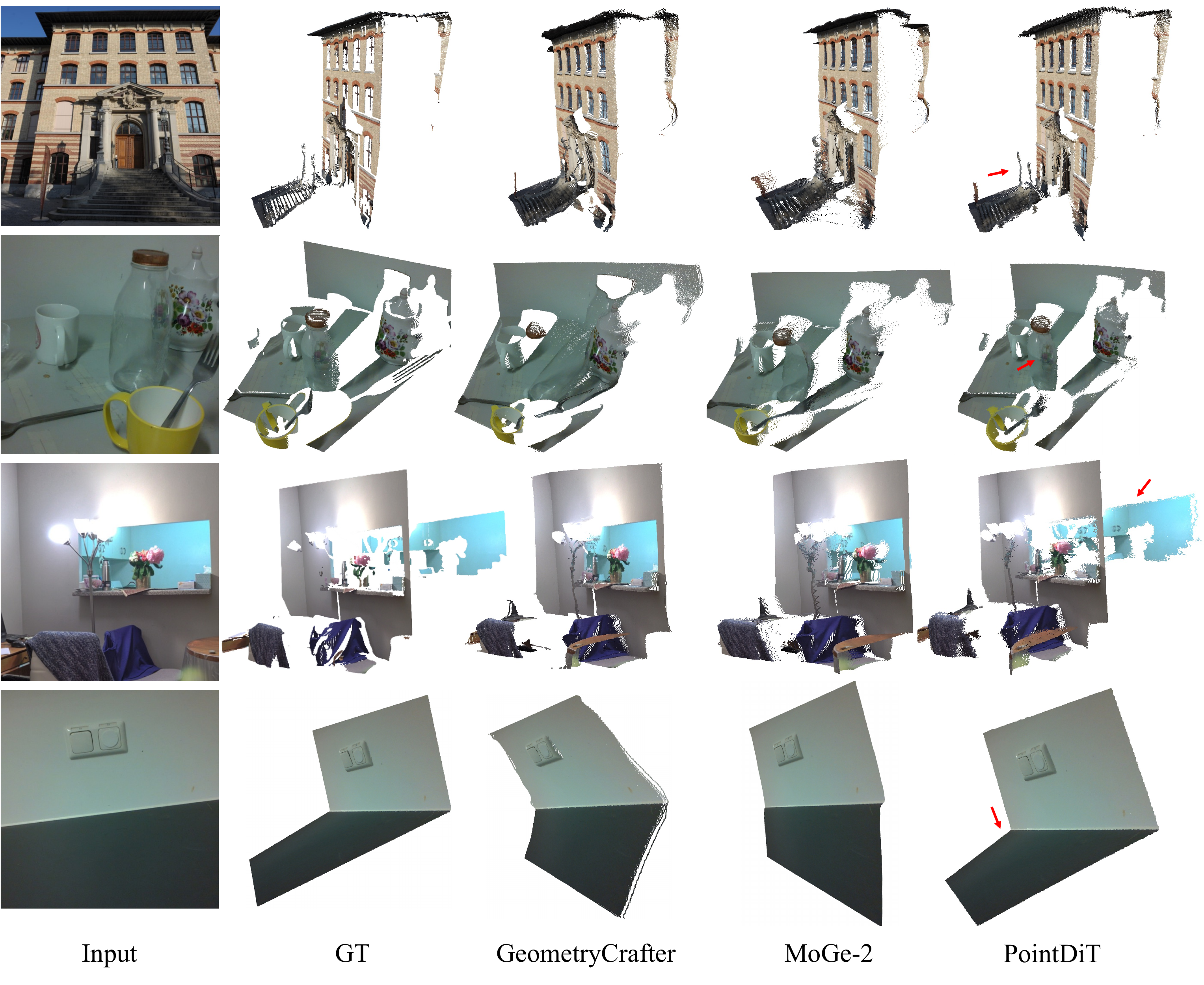}
    \caption{\textbf{Point map comparisons.} Our PointDiT is significantly better in terms of reconstructing thin structures (1st row), transparent objects (2nd rows), and maintaining a more accurate relative scale across the global scene (3rd and 4th rows). 
    We show additional depth comparisons in the appendix (\cref{fig:depth_compare}). 
    } 
    \label{fig:visual_compare}
\end{figure*}

\subsection{Evaluation Results}
\boldstartspace{Main Comparisons.} \cref{tab:comparison} reports average results over the seven evaluation datasets at $512 \times 512$. Our largest model, PointDiT-H, achieves the best depth accuracy (Rel$^\text{d}$ and $\delta_1^\text{d}$) and the best point map $\delta_1^\text{p}$, while PointDiT achieves the highest boundary sharpness (BF1) among all methods. The improvement is most pronounced on BF1: PointDiT raises boundary sharpness from $9.41$ (the best baseline) to $10.50$, reflecting markedly sharper geometry (\cref{fig:visual_compare}). On Rel$^\text{p}$, MoGe remains slightly ahead ($4.21$ \vs $4.40$), yet PointDiT is more accurate on every depth metric. PointDiT-L attains comparable boundary quality at lower cost, and our smallest variant, PointDiT-B, stays competitive with fewer parameters. Compared with PPD~\cite{xu2025pixel}, the most closely related pixel-space diffusion model, PointDiT is substantially better across all metrics and runs faster thanks to its fewer sampling steps. As PPD predicts only monocular depth, we compute its point map metrics by recovering camera intrinsics with MoGe-2~\cite{wang2025moge2}, following the official PPD repository.

\begin{figure*}[t]
    \centering
    \begin{subfigure}{0.48\linewidth}
        \centering
        \includegraphics[width=\linewidth]{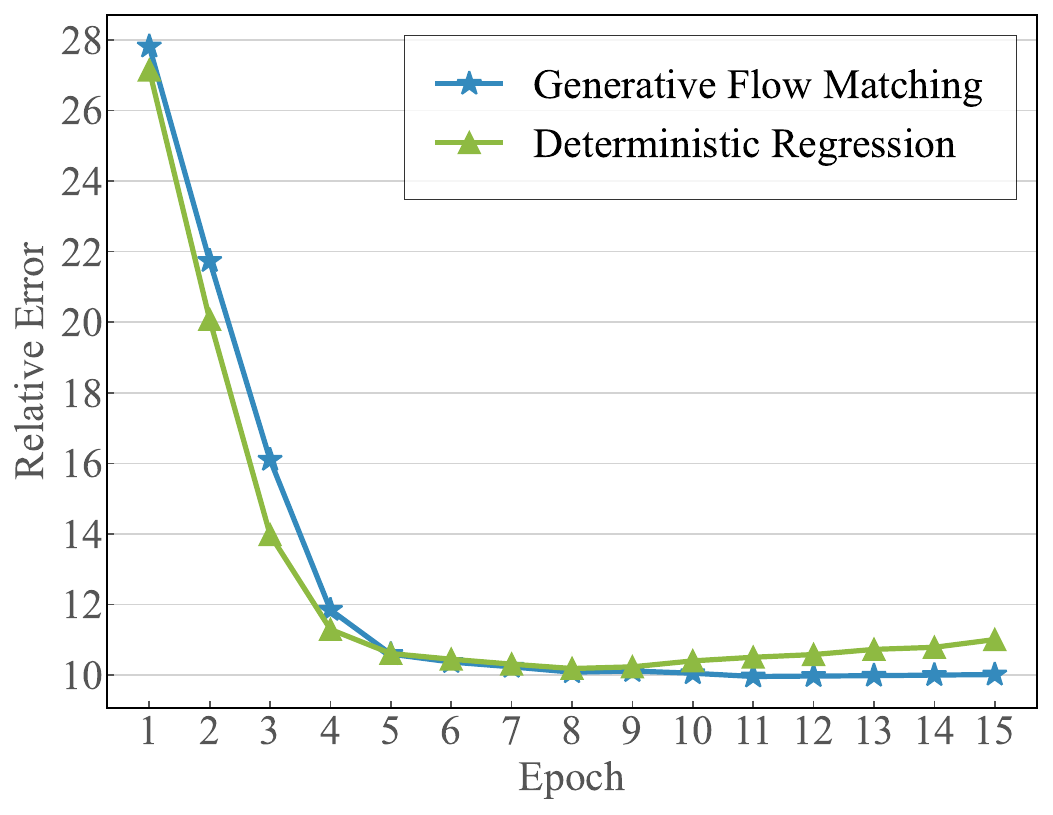}
        \caption{\textbf{Validation curves.}}
        \label{fig:vs_regression_curve}
    \end{subfigure}
    \hfill
    \begin{subfigure}{0.48\linewidth}
        \centering
        \includegraphics[width=\linewidth]{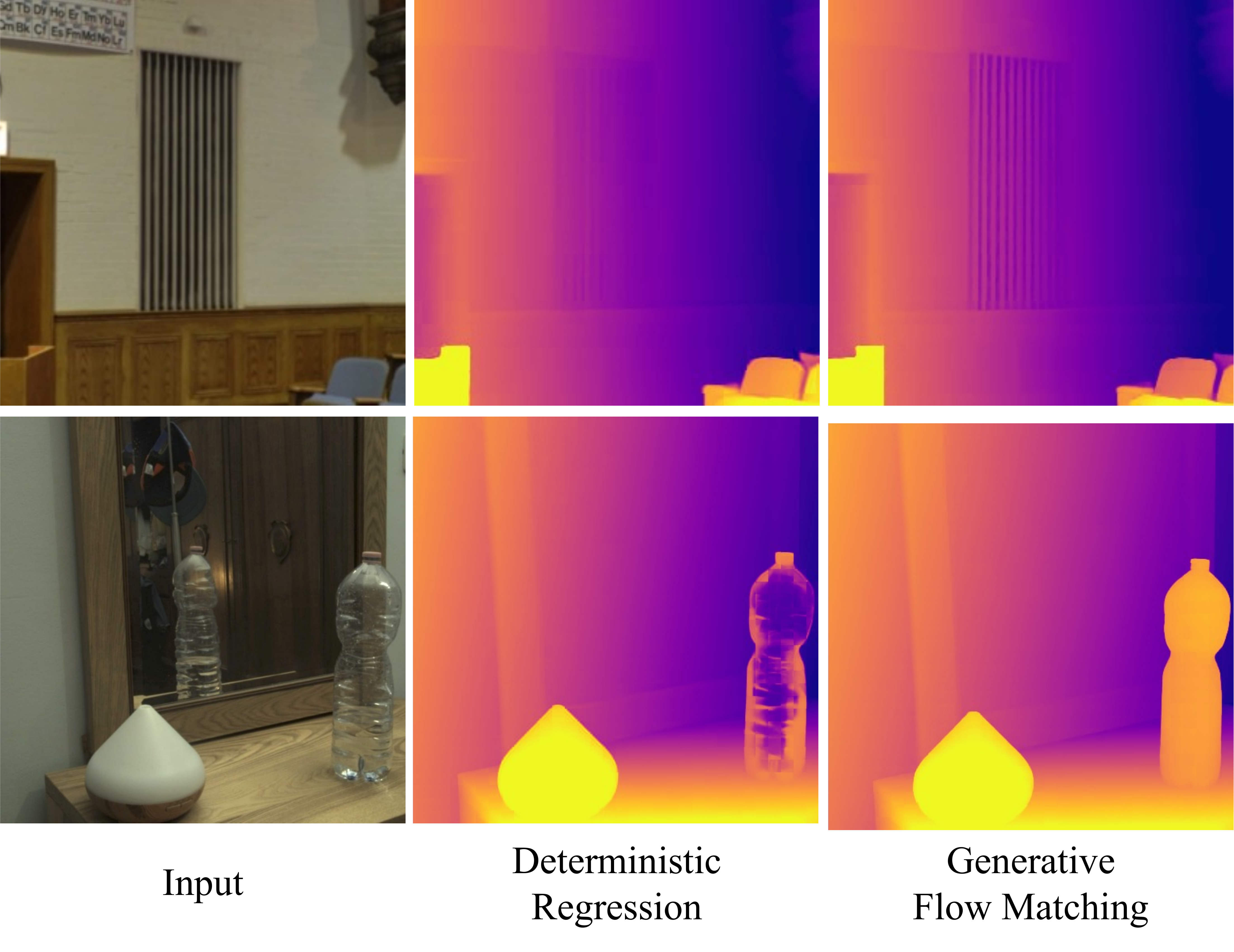}
        \caption{\textbf{Structural details and transparent objects.}}
        \label{fig:vs_regression_viz}
    \end{subfigure}
    \caption{\textbf{Generative flow matching \vs deterministic regression.} (a) The deterministic regressor converges faster at first but soon overfits, while the generative model trains stably and reaches lower error. (b) The generative model recovers sharper boundaries, thin structures, and transparent objects than the deterministic regressor. Overall, the generative formulation improves the boundary metric BF1 from $10.90$ to $13.92$ under this controlled comparison.}
    \label{fig:flowmatching_vs_regression}
\end{figure*}

\boldstartspace{Sharper Local Structures.} \cref{fig:visual_compare} shows qualitative comparisons with previous methods. PointDiT produces noticeably sharper local structures while preserving high-quality overall geometry. Unlike the latent-diffusion method GeometryCrafter, our approach avoids the lossy VAE compression that degrades fine detail, particularly at object boundaries (\cref{fig:vae_recon}). By removing the VAE entirely, PointDiT recovers substantially more accurate local structures, as reflected in the BF1 metric (\cref{tab:comparison}).

\boldstartspace{Single-Step Feed-Forward Inference.} Although our model is trained with flow matching, we find that it can perform single-step feed-forward inference, as also observed for diffusion-based monocular depth estimation~\cite{garcia2025fine}. In \cref{tab:infer_noise_sample}, we study the model's sensitivity to noise sampling in single-step inference, and find it highly robust across stochastic initializations. Across different random noise seeds, performance fluctuations are negligible, with $\text{Rel}^p$ and $\delta_1^p$ remaining nearly constant. More notably, the model maintains high-fidelity predictions even under deterministic sampling, where the input noise is set to all zeros. This ``all-zeros'' configuration not only matches but occasionally exceeds stochastic sampling. These results suggest that our model has learned a robust mapping from the DINOv3-encoded image patch tokens to the geometric point map. Remarkably, even with a single step, PointDiT-H already outperforms prior methods (\cref{tab:comparison}), at a fraction of the inference cost of latent diffusion models.

\boldstartspace{Multi-Step Refinement.} Thanks to its flow matching formulation, PointDiT can also benefit from additional inference steps using the same model. As shown in \cref{tab:comparison}, more steps steadily improve the boundary metric BF1, while Rel and $\delta_1$ remain stable, since a single step already yields high-quality results. \cref{fig:details_vs_steps} shows the corresponding visual refinement. Supporting a variable number of inference steps with one network underscores the flexibility of our approach.

\begin{table}[t]
\centering
\footnotesize
\caption{\textbf{Ablation experiments.} Trained on $256\times256$ SceneNet RGB-D and averaged over the seven unseen test sets with single-step inference (PointDiT-L). Our \colorbox{defaultcolor}{default} setting is highlighted in gray.}
\vspace{1mm}
\setlength{\tabcolsep}{3.5pt}
\begin{tabular}{lccccc}
\toprule
Setting & Rel$^\text{p}$ $\downarrow$ & $\delta_1^\text{p}$ $\uparrow$ & Rel$^\text{d}$ $\downarrow$ & $\delta_1^\text{d}$ $\uparrow$ & BF1 $\uparrow$ \\
\midrule
\multicolumn{6}{l}{\textit{(a) Prediction target}} \\
$v$-pred & 35.44 & 30.03 & 24.07 & 58.21 & 0.46 \\
\dft $x$-pred & \textbf{9.29} & \textbf{91.18} & \textbf{5.54} & \textbf{95.08} & \textbf{13.47} \\
\midrule
\multicolumn{6}{l}{\textit{(b) Noise schedule} ($t$-shift)} \\
$-0.8$ & 12.19 & 84.82 & 7.80 & 91.34 & 8.05 \\
$-1.2$ & 11.86 & 85.53 & 7.46 & 91.87 & \textbf{8.11} \\
$-1.6$ & 10.73 & 88.06 & 6.74 & 93.09 & 7.31 \\
\dft $-0.8$ \& rand zero & \textbf{9.68} & \textbf{90.54} & \textbf{6.00} & \textbf{94.64} & 7.24 \\
\midrule
\multicolumn{6}{l}{\textit{(c) Image patch embedding}} \\
Linear & 13.32 & 82.56 & 9.64 & 88.09 & 9.68 \\
DINOv2 (last layer) & 9.80 & 90.07 & 5.99 & 94.34 & 5.11 \\
DINOv3 (last layer) & 9.68 & 90.54 & 6.00 & 94.64 & 7.24 \\
\dft DINOv3 (4 layers) & 9.29 & 91.18 & 5.54 & 95.08 & \textbf{13.47} \\
\addlinespace[2pt]
MoGe-2 (4 layers) & 8.29 & \textbf{93.35} & 4.93 & 96.19 & 11.75 \\
DA3 (4 layers) & \textbf{8.26} & 93.09 & \textbf{4.74} & \textbf{96.47} & 12.58 \\
\midrule
\multicolumn{6}{l}{\textit{(d) Training loss}} \\
$v$-loss & 9.29 & 91.18 & 5.54 & \textbf{95.08} & 13.47 \\
\dft $v$-loss \& point loss & \textbf{9.10} & \textbf{91.48} & \textbf{5.53} & 94.88 & \textbf{13.92} \\
\midrule
\multicolumn{6}{l}{\textit{(e) Patch size} ($512\times512$)} \\
32 & 5.35 & 96.88 & 3.48 & 97.78 & 6.17 \\
\dft 16 & \textbf{5.01} & \textbf{97.34} & \textbf{3.06} & \textbf{98.17} & \textbf{10.37} \\
\bottomrule
\end{tabular}
\label{tab:ablation}
\end{table}

\subsection{Ablation and Analysis}

In this section, we evaluate the design choices of our model with controlled experiments. By default we train on the $256\times 256$ SceneNet-RGBD dataset and report the average metrics on the seven unseen test sets with single-step inference.

\boldstartspace{Generative Flow Matching \vs Deterministic Regression.} To further demonstrate the benefits of the generative flow matching formulation, we train a deterministic regressor by fixing both the time step and the noise to 0, while keeping exactly the same network architecture, training data, and training procedure. \cref{fig:vs_regression_curve} shows the validation curves of the two models. The deterministic regressor converges faster at first but soon overfits, whereas the generative model remains stable and ultimately reaches lower error. The gain in boundary quality is especially large, with a BF1 of $13.92$ (generative) \vs $10.90$ (deterministic). \cref{fig:vs_regression_viz} shows the corresponding visual comparison: the deterministic regressor produces blurry boundaries and fails to recover thin structures and transparent objects, while the generative model reconstructs much sharper boundaries and handles transparent objects more robustly. This controlled experiment confirms that the generative formulation learns sharper geometry and better handles ambiguous regions.

\boldstartspace{Prediction Target: $x$-Prediction \vs $v$-Prediction.} Key to our method's success is predicting the clean point map ($x$-prediction) rather than the velocity ($v$-prediction). \cref{tab:ablation}(a) shows that $v$-prediction fails catastrophically, consistent with the findings of JiT~\cite{li2025back} for image generation. We demonstrate that $x$-prediction is equally crucial for 3D point map generation.

\boldstartspace{Noise Schedule.} The original JiT for image generation uses a logit-normal noise schedule. However, we find that this alone yields unsatisfactory results for point map generation, where we measure per-point accuracy. Because the logit-normal sampler maps the timestep through a sigmoid, it is nearly impossible to draw an exact 0 during training. As a result, the model rarely sees pure noise, causing a train-test discrepancy that hurts performance~\cite{lin2024common}. Shifting the schedule toward high-noise regions with a smaller mean partially alleviates this, but the issue remains (\cref{tab:ablation}(b)). We instead randomly set the sampled timestep to exactly 0 with $10\%$ probability, a simple fix that substantially improves quality.

\boldstartspace{Image Patch Embeddings.} In \cref{tab:ablation}(c), we compare different patch embedding methods. Even without any pre-trained image backbone (\ie, with plain linear embeddings), our model already achieves decent results. Pre-trained embeddings nonetheless help: comparing the last-layer features of DINOv2 and DINOv3, DINOv3 performs slightly better. Uniformly sampling 4 intermediate layers improves the results further, especially the BF1 metric, indicating the benefit of integrating different levels of abstraction from pre-trained backbones. We additionally evaluate the embeddings of MoGe-2~\cite{wang2025moge2} and DA3~\cite{lin2025depth}, which are specifically fine-tuned for monocular geometry estimation. They further improve the accuracy metrics (Rel and $\delta_1$), confirming that our model readily benefits from geometry-aware features. Interestingly, however, DINOv3 still attains the best boundary sharpness (BF1 of $13.47$ \vs $11.75$ for MoGe-2 and $12.58$ for DA3). We hypothesize that, because MoGe-2 and DA3 are trained with regression objectives that tend to over-smooth geometry, their representations retain less high-frequency boundary information, despite encoding more accurate global structure. We do not, however, use any of these features in our main results, as our focus is to demonstrate the effectiveness of the pure pixel-space diffusion framework. Since both MoGe-2 and DA3 are themselves fine-tuned from pre-trained DINO features, we deliberately keep the same pre-trained model for a controlled comparison, so that our improvements can be attributed to the framework itself rather than to stronger, task-specific features.

\boldstartspace{Training Loss.} The ablation results discussed so far use only the flow matching loss (\cref{eq:fm_loss}), which is already highly effective at recovering high-quality geometry. Adding the relative loss (\cref{eq:rel_point}), specifically designed for point map data, further improves the results, as shown in \cref{tab:ablation}(d).

\begin{figure}[t]
    \centering
    \includegraphics[width=\linewidth]{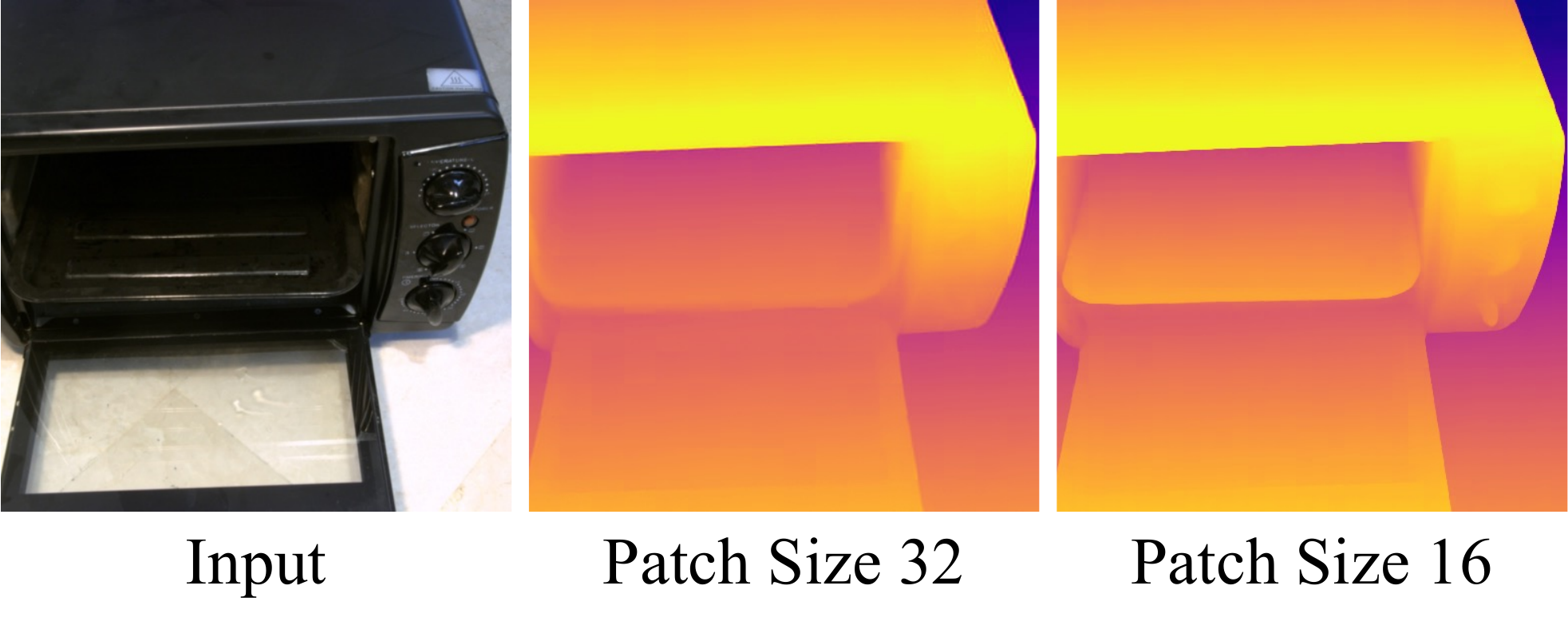}
    \caption{\textbf{Effect of patch size.} At $512\times512$ resolution, a patch size of 16 recovers sharper boundaries and finer local structures than a patch size of 32.}
    \label{fig:patch_size}
\end{figure}

\boldstartspace{Patch Size.} We further evaluate the impact of patch size on high-resolution images by fine-tuning the $256\times 256$ pre-trained model to $512 \times 512$ resolution. To save compute, the $512\times 512$ models in this part are fine-tuned on a 6-dataset subset (Hypersim, VKITTI2, UrbanSyn, Synscapes, TartanAir, and OmniWorldGame; $1.48$M samples) rather than the full training set. Comparing patch sizes 16 and 32, we find that 16 yields better overall results and sharper boundaries (\cref{tab:ablation}(e)). This is expected, since point map prediction requires pixel-perfect accuracy, and a larger patch size tends to discard more high-frequency detail. \cref{fig:patch_size} shows the corresponding qualitative comparison.

\section{Conclusion}

We presented a minimalist pixel-space diffusion model for monocular point map prediction that removes the architectural overhead of VAEs and hybrid networks. A plain Vision Transformer trained directly on raw point maps, conditioned on DINOv3 features, already produces accurate geometry with notably sharper boundaries and supports both single-step and multi-step inference. By showing that dense geometry can be modeled effectively in pixel space, we bridge the gap between standard image generation and 3D reconstruction, paving the way for VAE-free, end-to-end 3D and 4D generation that relies solely on direct diffusion to model complex structural distributions.

\boldstartspace{Limitation.} While our framework delivers robust geometric estimation, it is currently trained at fixed resolutions ($256\times 256$ and $512\times 512$); mixed-resolution training is a promising direction for generalizing seamlessly across image resolutions. In addition, our model still has room for improvement on outdoor scenes (\cref{tab:per_dataset_point}) due to the relatively limited scale and diversity of the training data. Further scaling of the training set, particularly with more outdoor data, will likely be key to realizing the model's full potential. Finally, our model currently predicts geometry alone. However, since the backbone is a plain ViT, extending it to jointly output additional modalities, such as RGB appearance, would be straightforward and require minimal architectural changes. The same flexibility makes it natural to explore multi-view generation, alternative 3D representations, and richer conditioning signals (\eg, camera parameters), which we view as exciting avenues for future work.

\boldstartspace{Acknowledgements.}
We thank Nando Metzger, Weirong Chen, Felix Wimbauer, Haiwen Huang, Gene Chou, Luca Zanella, Erik Sandström, Keisuke Tateno, Goutam Bhat, Mattia Segu, Vasile Lup, Tobias Fischer, Shaohui Liu and Bingxin Ke for the insightful discussions and support.

\section*{Impact Statement}
This paper presents work whose goal is to advance the field of machine learning and computer vision by simplifying the paradigm for monocular 3D reconstruction. There are some potential societal consequences of our work, ranging from advancements in robotics to spatial intelligence, none of which we feel must be specifically highlighted here from an ethical standpoint.

\bibliography{example_paper}

\begin{thebibliography}{50}
\providecommand{\natexlab}[1]{#1}
\providecommand{\url}[1]{\texttt{#1}}
\expandafter\ifx\csname urlstyle\endcsname\relax
  \providecommand{\doi}[1]{doi: #1}\else
  \providecommand{\doi}{doi: \begingroup \urlstyle{rm}\Url}\fi

\bibitem[{Black Forest Labs}(2025)]{bfl2025representation}
{Black Forest Labs}.
\newblock {FLUX.2}: Analyzing and enhancing the latent space of {FLUX} -- representation comparison, 2025.
\newblock URL \url{https://bfl.ai/research/representation-comparison}.

\bibitem[Bochkovskii et~al.(2025)Bochkovskii, Delaunoy, Germain, Santos, Zhou, Richter, and Koltun]{bochkovskii2025depth}
Bochkovskii, A., Delaunoy, A., Germain, H., Santos, M., Zhou, Y., Richter, S.~R., and Koltun, V.
\newblock Depth pro: Sharp monocular metric depth in less than a second.
\newblock In \emph{ICLR}, 2025.

\bibitem[Cabon et~al.(2020)Cabon, Murray, and Humenberger]{cabon2020virtual}
Cabon, Y., Murray, N., and Humenberger, M.
\newblock Virtual kitti 2.
\newblock \emph{arXiv preprint arXiv:2001.10773}, 2020.

\bibitem[Dosovitskiy(2020)]{dosovitskiy2020image}
Dosovitskiy, A.
\newblock An image is worth 16x16 words: Transformers for image recognition at scale.
\newblock \emph{arXiv preprint arXiv:2010.11929}, 2020.

\bibitem[Eigen et~al.(2014)Eigen, Puhrsch, and Fergus]{eigen2014depth}
Eigen, D., Puhrsch, C., and Fergus, R.
\newblock Depth map prediction from a single image using a multi-scale deep network.
\newblock \emph{NIPS}, 27, 2014.

\bibitem[Esser et~al.(2024)Esser, Kulal, Blattmann, Entezari, M{\"u}ller, Saini, Levi, Lorenz, Sauer, Boesel, et~al.]{esser2024scaling}
Esser, P., Kulal, S., Blattmann, A., Entezari, R., M{\"u}ller, J., Saini, H., Levi, Y., Lorenz, D., Sauer, A., Boesel, F., et~al.
\newblock Scaling rectified flow transformers for high-resolution image synthesis.
\newblock In \emph{ICML}, 2024.

\bibitem[Garcia et~al.(2025)Garcia, Zeid, Schmidt, De~Geus, Hermans, and Leibe]{garcia2025fine}
Garcia, G.~M., Zeid, K.~A., Schmidt, C., De~Geus, D., Hermans, A., and Leibe, B.
\newblock Fine-tuning image-conditional diffusion models is easier than you think.
\newblock In \emph{WACV}, 2025.

\bibitem[Geiger et~al.(2012)Geiger, Lenz, and Urtasun]{geiger2012kitti}
Geiger, A., Lenz, P., and Urtasun, R.
\newblock Are we ready for autonomous driving? the kitti vision benchmark suite.
\newblock In \emph{CVPR}, 2012.

\bibitem[G{\'o}mez et~al.(2025)G{\'o}mez, Silva, Seoane, Borr{\'a}s, Noriega, Ros, Iglesias-Guitian, and L{\'o}pez]{gomez2025all}
G{\'o}mez, J.~L., Silva, M., Seoane, A., Borr{\'a}s, A., Noriega, M., Ros, G., Iglesias-Guitian, J.~A., and L{\'o}pez, A.~M.
\newblock All for one, and one for all: Urbansyn dataset, the third musketeer of synthetic driving scenes.
\newblock \emph{Neurocomputing}, 637:\penalty0 130038, 2025.

\bibitem[He et~al.(2024)He, Li, Yin, Liang, Li, Zhou, Zhang, Liu, and Chen]{he2024lotus}
He, J., Li, H., Yin, W., Liang, Y., Li, L., Zhou, K., Zhang, H., Liu, B., and Chen, Y.-C.
\newblock Lotus: Diffusion-based visual foundation model for high-quality dense prediction.
\newblock \emph{arXiv preprint arXiv:2409.18124}, 2024.

\bibitem[Ho et~al.(2020)Ho, Jain, and Abbeel]{ho2020denoising}
Ho, J., Jain, A., and Abbeel, P.
\newblock Denoising diffusion probabilistic models.
\newblock \emph{NIPS}, 33, 2020.

\bibitem[Hu et~al.(2025)Hu, Gao, Li, Zhao, Cun, Zhang, Quan, and Shan]{hu2025depthcrafter}
Hu, W., Gao, X., Li, X., Zhao, S., Cun, X., Zhang, Y., Quan, L., and Shan, Y.
\newblock Depthcrafter: Generating consistent long depth sequences for open-world videos.
\newblock In \emph{CVPR}, 2025.

\bibitem[Huang et~al.(2018)Huang, Matzen, Kopf, Ahuja, and Huang]{huang2018deepmvs}
Huang, P.-H., Matzen, K., Kopf, J., Ahuja, N., and Huang, J.-B.
\newblock Deepmvs: Learning multi-view stereopsis.
\newblock In \emph{CVPR}, 2018.

\bibitem[Jung et~al.(2022)Jung, Ruhkamp, Zhai, Brasch, Li, Verdie, Song, Zhou, Armagan, Ilic, et~al.]{jung2022hammer}
Jung, H., Ruhkamp, P., Zhai, G., Brasch, N., Li, Y., Verdie, Y., Song, J., Zhou, Y., Armagan, A., Ilic, S., et~al.
\newblock Is my depth ground-truth good enough? hammer -- highly accurate multi-modal dataset for dense 3d scene regression.
\newblock \emph{arXiv preprint arXiv:2205.04565}, 2022.

\bibitem[Karaev et~al.(2023)Karaev, Rocco, Graham, Neverova, Vedaldi, and Rupprecht]{karaev2023dynamicstereo}
Karaev, N., Rocco, I., Graham, B., Neverova, N., Vedaldi, A., and Rupprecht, C.
\newblock Dynamicstereo: Consistent dynamic depth from stereo videos.
\newblock In \emph{CVPR}, 2023.

\bibitem[Ke et~al.(2024)Ke, Obukhov, Huang, Metzger, Daudt, and Schindler]{ke2024repurposing}
Ke, B., Obukhov, A., Huang, S., Metzger, N., Daudt, R.~C., and Schindler, K.
\newblock Repurposing diffusion-based image generators for monocular depth estimation.
\newblock In \emph{CVPR}, 2024.

\bibitem[Kingma \& Welling(2014)Kingma and Welling]{kingma2013auto}
Kingma, D.~P. and Welling, M.
\newblock Auto-encoding variational bayes.
\newblock 2014.

\bibitem[Koch et~al.(2018)Koch, Liebel, Fraundorfer, and K{\"o}rner]{koch2018ibims}
Koch, T., Liebel, L., Fraundorfer, F., and K{\"o}rner, M.
\newblock Evaluation of cnn-based single-image depth estimation methods.
\newblock In \emph{ECCV Workshops}, 2018.

\bibitem[L{\^e} et~al.(2021)L{\^e}, Mensink, Das, and Gevers]{le2021eden}
L{\^e}, H.-{\^A}., Mensink, T., Das, P., and Gevers, T.
\newblock Eden: Multimodal synthetic dataset of enclosed garden scenes.
\newblock In \emph{WACV}, 2021.

\bibitem[Li \& He(2026)Li and He]{li2025back}
Li, T. and He, K.
\newblock Back to basics: Let denoising generative models denoise.
\newblock In \emph{CVPR}, 2026.

\bibitem[Lin et~al.(2026)Lin, Chen, Liew, Chen, Li, Shi, Feng, and Kang]{lin2025depth}
Lin, H., Chen, S., Liew, J., Chen, D.~Y., Li, Z., Shi, G., Feng, J., and Kang, B.
\newblock Depth anything 3: Recovering the visual space from any views.
\newblock In \emph{ICLR}, 2026.

\bibitem[Lin et~al.(2024)Lin, Liu, Li, and Yang]{lin2024common}
Lin, S., Liu, B., Li, J., and Yang, X.
\newblock Common diffusion noise schedules and sample steps are flawed.
\newblock In \emph{WACV}, 2024.

\bibitem[Loshchilov \& Hutter(2019)Loshchilov and Hutter]{loshchilov2017decoupled}
Loshchilov, I. and Hutter, F.
\newblock Decoupled weight decay regularization.
\newblock 2019.

\bibitem[McCormac et~al.(2017)McCormac, Handa, Leutenegger, and Davison]{mccormac2017scenenet}
McCormac, J., Handa, A., Leutenegger, S., and Davison, A.~J.
\newblock Scenenet rgb-d: Can 5m synthetic images beat generic imagenet pre-training on indoor segmentation?
\newblock In \emph{ICCV}, 2017.

\bibitem[Peebles \& Xie(2023)Peebles and Xie]{peebles2023scalable}
Peebles, W. and Xie, S.
\newblock Scalable diffusion models with transformers.
\newblock In \emph{ICCV}, 2023.

\bibitem[Piccinelli et~al.(2025)Piccinelli, Sakaridis, Yang, Segu, Li, Abbeloos, and Van~Gool]{piccinelli2025unidepthv2}
Piccinelli, L., Sakaridis, C., Yang, Y.-H., Segu, M., Li, S., Abbeloos, W., and Van~Gool, L.
\newblock Unidepthv2: Universal monocular metric depth estimation made simpler.
\newblock \emph{TPAMI}, 2025.

\bibitem[Ranftl et~al.(2021)Ranftl, Bochkovskiy, and Koltun]{ranftl2021vision}
Ranftl, R., Bochkovskiy, A., and Koltun, V.
\newblock Vision transformers for dense prediction.
\newblock In \emph{ICCV}, 2021.

\bibitem[Roberts et~al.(2021)Roberts, Ramapuram, Ranjan, Kumar, Bautista, Paczan, Webb, and Susskind]{roberts2021hypersim}
Roberts, M., Ramapuram, J., Ranjan, A., Kumar, A., Bautista, M.~A., Paczan, N., Webb, R., and Susskind, J.~M.
\newblock Hypersim: A photorealistic synthetic dataset for holistic indoor scene understanding.
\newblock In \emph{ICCV}, 2021.

\bibitem[Rombach et~al.(2022)Rombach, Blattmann, Lorenz, Esser, and Ommer]{rombach2022high}
Rombach, R., Blattmann, A., Lorenz, D., Esser, P., and Ommer, B.
\newblock High-resolution image synthesis with latent diffusion models.
\newblock In \emph{CVPR}, 2022.

\bibitem[Salimans \& Ho(2022)Salimans and Ho]{salimans2022progressive}
Salimans, T. and Ho, J.
\newblock Progressive distillation for fast sampling of diffusion models.
\newblock \emph{arXiv preprint arXiv:2202.00512}, 2022.

\bibitem[Sch{\"o}ps et~al.(2017)Sch{\"o}ps, Sch{\"o}nberger, Galliani, Sattler, Schindler, Pollefeys, and Geiger]{schops2017eth3d}
Sch{\"o}ps, T., Sch{\"o}nberger, J.~L., Galliani, S., Sattler, T., Schindler, K., Pollefeys, M., and Geiger, A.
\newblock A multi-view stereo benchmark with high-resolution images and multi-camera videos.
\newblock In \emph{CVPR}, 2017.

\bibitem[Silberman et~al.(2012)Silberman, Hoiem, Kohli, and Fergus]{silberman2012nyuv2}
Silberman, N., Hoiem, D., Kohli, P., and Fergus, R.
\newblock Indoor segmentation and support inference from rgbd images.
\newblock In \emph{ECCV}, 2012.

\bibitem[Sim{\'e}oni et~al.(2025)Sim{\'e}oni, Vo, Seitzer, Baldassarre, Oquab, Jose, Khalidov, Szafraniec, Yi, Ramamonjisoa, et~al.]{simeoni2025dinov3}
Sim{\'e}oni, O., Vo, H.~V., Seitzer, M., Baldassarre, F., Oquab, M., Jose, C., Khalidov, V., Szafraniec, M., Yi, S., Ramamonjisoa, M., et~al.
\newblock Dinov3.
\newblock \emph{arXiv preprint arXiv:2508.10104}, 2025.

\bibitem[Szymanowicz et~al.(2025)Szymanowicz, Zhang, Srinivasan, Gao, Brussee, Holynski, Martin-Brualla, Barron, and Henzler]{szymanowicz2025bolt3d}
Szymanowicz, S., Zhang, J.~Y., Srinivasan, P., Gao, R., Brussee, A., Holynski, A., Martin-Brualla, R., Barron, J.~T., and Henzler, P.
\newblock Bolt3d: Generating 3d scenes in seconds.
\newblock In \emph{ICCV}, 2025.

\bibitem[Vasiljevic et~al.(2019)Vasiljevic, Kolkin, Zhang, Luo, Wang, Dai, Daniele, Mostajabi, Basart, Walter, and Shakhnarovich]{vasiljevic2019diode}
Vasiljevic, I., Kolkin, N., Zhang, S., Luo, R., Wang, H., Dai, F.~Z., Daniele, A.~F., Mostajabi, M., Basart, S., Walter, M.~R., and Shakhnarovich, G.
\newblock Diode: A dense indoor and outdoor depth dataset.
\newblock \emph{arXiv preprint arXiv:1908.00463}, 2019.

\bibitem[Wang et~al.(2025{\natexlab{a}})Wang, Chen, Karaev, Vedaldi, Rupprecht, and Novotny]{wang2025vggt}
Wang, J., Chen, M., Karaev, N., Vedaldi, A., Rupprecht, C., and Novotny, D.
\newblock Vggt: Visual geometry grounded transformer.
\newblock In \emph{CVPR}, 2025{\natexlab{a}}.

\bibitem[Wang et~al.(2019)Wang, Zheng, Yan, Deng, Zhao, and Chu]{wang2019irs}
Wang, Q., Zheng, S., Yan, Q., Deng, F., Zhao, K., and Chu, X.
\newblock Irs: A large synthetic indoor robotics stereo dataset for disparity and surface normal estimation.
\newblock \emph{arXiv preprint arXiv:1912.09632}, 2019.

\bibitem[Wang et~al.(2025{\natexlab{b}})Wang, Xu, Dai, Xiang, Deng, Tong, and Yang]{wang2025moge}
Wang, R., Xu, S., Dai, C., Xiang, J., Deng, Y., Tong, X., and Yang, J.
\newblock Moge: Unlocking accurate monocular geometry estimation for open-domain images with optimal training supervision.
\newblock In \emph{CVPR}, 2025{\natexlab{b}}.

\bibitem[Wang et~al.(2025{\natexlab{c}})Wang, Xu, Dong, Deng, Xiang, Lv, Sun, Tong, and Yang]{wang2025moge2}
Wang, R., Xu, S., Dong, Y., Deng, Y., Xiang, J., Lv, Z., Sun, G., Tong, X., and Yang, J.
\newblock Moge-2: Accurate monocular geometry with metric scale and sharp details.
\newblock \emph{arXiv preprint arXiv:2507.02546}, 2025{\natexlab{c}}.

\bibitem[Wang et~al.(2020)Wang, Zhu, Wang, Hu, Qiu, Wang, Hu, Kapoor, and Scherer]{wang2020tartanair}
Wang, W., Zhu, D., Wang, X., Hu, Y., Qiu, Y., Wang, C., Hu, Y., Kapoor, A., and Scherer, S.
\newblock Tartanair: A dataset to push the limits of visual slam.
\newblock In \emph{IEEE/RSJ International Conference on Intelligent Robots and Systems (IROS)}, 2020.

\bibitem[Wang et~al.(2025{\natexlab{d}})]{wang2025tartanground}
Wang, W. et~al.
\newblock Tartanground: A large-scale dataset for ground robot perception and navigation.
\newblock \emph{arXiv preprint arXiv:2505.10696}, 2025{\natexlab{d}}.

\bibitem[Wrenninge \& Unger(2018)Wrenninge and Unger]{wrenninge2018synscapes}
Wrenninge, M. and Unger, J.
\newblock Synscapes: A photorealistic synthetic dataset for street scene parsing.
\newblock \emph{arXiv preprint arXiv:1810.08705}, 2018.

\bibitem[Xu et~al.(2025{\natexlab{a}})Xu, Lin, Luo, Wang, Yao, Zhu, Pu, Chi, Sun, Wang, et~al.]{xu2025pixel}
Xu, G., Lin, H., Luo, H., Wang, X., Yao, J., Zhu, L., Pu, Y., Chi, C., Sun, H., Wang, B., et~al.
\newblock Pixel-perfect depth with semantics-prompted diffusion transformers.
\newblock In \emph{NeurIPS}, 2025{\natexlab{a}}.

\bibitem[Xu et~al.(2025{\natexlab{b}})Xu, Gao, Hu, Li, Zhang, and Shan]{xu2025geometrycrafter}
Xu, T.-X., Gao, X., Hu, W., Li, X., Zhang, S.-H., and Shan, Y.
\newblock Geometrycrafter: Consistent geometry estimation for open-world videos with diffusion priors.
\newblock In \emph{ICCV}, 2025{\natexlab{b}}.

\bibitem[Yang et~al.(2024)Yang, Kang, Huang, Xu, Feng, and Zhao]{yang2024depth}
Yang, L., Kang, B., Huang, Z., Xu, X., Feng, J., and Zhao, H.
\newblock Depth anything: Unleashing the power of large-scale unlabeled data.
\newblock In \emph{CVPR}, 2024.

\bibitem[Yao et~al.(2025)Yao, Yang, and Wang]{yao2025reconstruction}
Yao, J., Yang, B., and Wang, X.
\newblock Reconstruction vs. generation: Taming optimization dilemma in latent diffusion models.
\newblock In \emph{CVPR}, 2025.

\bibitem[Yu et~al.(2025)Yu, Kwak, Jang, Jeong, Huang, Shin, and Xie]{yu2025representation}
Yu, S., Kwak, S., Jang, H., Jeong, J., Huang, J., Shin, J., and Xie, S.
\newblock Representation alignment for generation: Training diffusion transformers is easier than you think.
\newblock 2025.

\bibitem[Zama~Ramirez et~al.(2022)Zama~Ramirez, Tosi, Poggi, Salti, Di~Stefano, and Mattoccia]{zamaramirez2022booster}
Zama~Ramirez, P., Tosi, F., Poggi, M., Salti, S., Di~Stefano, L., and Mattoccia, S.
\newblock Open challenges in deep stereo: The booster dataset.
\newblock In \emph{CVPR}, 2022.

\bibitem[Zheng et~al.(2025)Zheng, Ma, Tong, and Xie]{zheng2025diffusion}
Zheng, B., Ma, N., Tong, S., and Xie, S.
\newblock Diffusion transformers with representation autoencoders.
\newblock \emph{arXiv preprint arXiv:2510.11690}, 2025.

\bibitem[Zhou et~al.(2025)]{zhou2025omniworld}
Zhou, Y. et~al.
\newblock Omniworld: A multi-domain and multi-modal dataset for 4d world modeling.
\newblock \emph{arXiv preprint arXiv:2509.12201}, 2025.

\end{thebibliography}
\bibliographystyle{icml2026}

\newpage
\appendix
\onecolumn

{\LARGE \bfseries Appendix}

\vspace{1em}

\section{Experimental Details}
\label{sec:exp_details}

We train our PointDiT on synthetic datasets that provide
dense, accurate ground-truth depth together with known camera
intrinsics, and we evaluate zero-shot on unseen real-world benchmarks. From each image we back-project the
depth map through the intrinsics into a per-pixel 3D point map,
which is the prediction target of the model. Training proceeds in two stages: a $256\times 256$\ pre-training
stage on a single large indoor dataset, followed by a $512\times 512$\
fine-tuning stage on a diverse multi-dataset mixture.

\subsection{Training Datasets}
\label{sec:train_data}

Table~\ref{tab:datasets} lists the training data of both stages.

\paragraph{Stage 1 (Pre-Training, $256\times256$).}
We first pre-train on SceneNet-RGBD~\cite{mccormac2017scenenet}, a single large-scale synthetic indoor
dataset ($\approx\!5.36$M samples). Its scale and
clean indoor geometry let the model cheaply acquire a strong image-to-point
prior at low resolution.

\paragraph{Stage 2 (Fine-Tuning, $512\times512$).}
We then fine-tune at $512\times512$ on a mixture of $11$ synthetic datasets
($\approx\!6.22$M samples) spanning indoor, outdoor ground-level, and
aerial/diverse domains, which adds the outdoor, large-scale, and high-detail
geometry absent from Stage~1. We combine these datasets through weighted
sampling. Each dataset $d$ is assigned a mixing
weight $w_d$ (Table~\ref{tab:datasets}, $\sum_d w_d{=}1$); every sample of dataset
$d$ receives the per-sample probability $w_d/N_d$, where $N_d$ is the number of
samples in $d$. After global normalization, the probability that a drawn sample
comes from dataset $d$ is therefore exactly $w_d$, independent of the
corpus size $N_d$.
This decouples the
effective data mixture from the highly imbalanced raw corpus sizes: \eg
TartanGround accounts for $67.1\%$ of all samples but is sampled only $15\%$ of
the time, while small high-quality sets such as Synscapes ($25$k samples) are
upsampled to $9\%$.

\subsection{Evaluation Datasets}
\label{sec:eval_data}

We evaluate zero-shot on seven real-world depth benchmarks, none of which
overlaps the (synthetic) training datasets of either stage; this measures
synthetic-to-real generalization. We report the
full-set sample counts in Table~\ref{tab:eval_datasets}. They span indoor,
outdoor, and mixed scenes, including three challenging boundary-focused sets with
transparent/specular surfaces and sharp planar discontinuities (HAMMER, iBims-1,
Booster), on which we additionally measure boundary sharpness.

\subsection{Training Details}
\label{sec:train_details}

\paragraph{Data Augmentation.}
We apply two groups of augmentations: geometric operations, applied
jointly to the image and the point map (with intrinsics updated accordingly), and
photometric operations, applied to the image only.
\begin{itemize}[itemsep=2pt, topsep=2pt, parsep=0pt, leftmargin=*]
    \item \textbf{Geometric} (image and point map):
    \begin{enumerate}[label=(\alph*), itemsep=1pt, topsep=1pt, parsep=0pt, leftmargin=1.8em]
        \item Resize to the working resolution. In Stage~2 this is aspect-ratio-preserving so the image height is $512$ (smaller images are upscaled; larger images are downscaled with probability $0.5$, otherwise cropped at native resolution); in Stage~1 the image height is set to $256$.
        \item A square random crop to the working resolution (center crop at test time).
        \item A random horizontal flip ($p{=}0.5$), which also negates the point $X$-coordinate to keep the geometry consistent with the flipped image.
    \end{enumerate}
    \item \textbf{Photometric} (image only): color jitter (brightness/contrast/saturation $0.1$, hue $0.05$) applied to every sample, plus appearance augmentations each applied independently with probability $0.2$: Gaussian blur, autocontrast, histogram equalization, random channel permutation, JPEG compression ($q\in[40,100]$), and grayscale conversion.
\end{itemize}

\paragraph{Optimization.}
Both stages use AdamW ($\beta_1{=}0.9$, $\beta_2{=}0.95$, weight decay $0$) in
\texttt{bf16} mixed precision, the linear learning-rate scaling rule
$\text{lr}=\text{blr}\cdot(\text{global batch})/256$, no gradient clipping or
accumulation, and two exponential moving averages of the weights (decays $0.9999$
and $0.9996$); the $0.9999$ EMA is used for all evaluations. Stage~1
($256\times256$) uses $\text{blr}{=}5\!\times\!10^{-5}$, a $5$-epoch linear warmup followed by a constant
learning rate, for $30$ epochs.
{Stage~2} ($512\times512$) is initialized from the Stage-1 checkpoint (the
fixed positional embeddings are bicubically interpolated from the $16\times16$ to
the $32\times32$ grid) and uses $\text{blr}{=}1\!\times\!10^{-4}$, a constant learning rate (no warmup). 
We report the per-variant epoch counts, GPU counts, and
wall-clock times for both stages in \cref{tab:training}.

\begin{table*}[t]
\centering
\small
\setlength{\tabcolsep}{4pt}
\caption{\textbf{Training datasets.} All sources are synthetic and provide dense depth with known camera intrinsics. Weight is the Stage-2 dataset mixing (sampling) probability, applied
independently of the corpus size. Stage~1 pre-trains on a single dataset.}
\label{tab:datasets}
\begin{tabular}{llrcc}
\toprule
Dataset & Domain & \#Samples & Weight & Website \\
\midrule
\multicolumn{5}{l}{Stage 1 --- pre-training ($256\times256$)} \\
SceneNet-RGBD~\cite{mccormac2017scenenet} & indoor & $\mathbf{5{,}359{,}500}$ & $1.00$ & \href{https://robotvault.bitbucket.io/scenenet-rgbd.html}{link} \\
\midrule
\multicolumn{5}{l}{Stage 2 --- fine-tuning ($512\times512$) mixture} \\
Hypersim~\cite{roberts2021hypersim} & indoor & $70{,}647$ & $0.12$ & \href{https://github.com/apple/ml-hypersim}{link} \\
Virtual KITTI 2~\cite{cabon2020virtual} & driving & $42{,}520$ & $0.14$ & \href{https://europe.naverlabs.com/research/proxy-virtual-worlds-vkitti-2/}{link} \\
UrbanSyn~\cite{gomez2025all} & urban driving & $7{,}539$ & $0.05$ & \href{https://www.urbansyn.org}{link} \\
Synscapes~\cite{wrenninge2018synscapes} & urban driving & $25{,}000$ & $0.09$ & \href{https://synscapes.on.liu.se}{link} \\
TartanAir~\cite{wang2020tartanair} & diverse & $306{,}637$ & $0.10$ & \href{https://theairlab.org/tartanair-dataset/}{link} \\
OmniWorld-Game~\cite{zhou2025omniworld} & diverse (game) & $1{,}024{,}252$ & $0.19$ & \href{https://huggingface.co/datasets/InternRobotics/OmniWorld}{link} \\
EDEN~\cite{le2021eden} & garden/outdoor & $368{,}663$ & $0.05$ & \href{https://lhoangan.github.io/eden/}{link} \\
IRS~\cite{wang2019irs} & indoor & $39{,}342$ & $0.02$ & \href{https://github.com/HKBU-HPML/IRS}{link} \\
Dynamic Replica~\cite{karaev2023dynamicstereo} & indoor & $150{,}900$ & $0.03$ & \href{https://dynamic-stereo.github.io}{link} \\
MVS-Synth~\cite{huang2018deepmvs} & urban & $12{,}000$ & $0.06$ & \href{https://phuang17.github.io/DeepMVS/mvs-synth.html}{link} \\
TartanGround~\cite{wang2025tartanground} & diverse & $4{,}170{,}178$ & $0.15$ & \href{https://tartanair.org/tartanground/index.html}{link} \\
\cmidrule(l){1-5}
Total (Stage 2) & & $\mathbf{6{,}217{,}678}$ & $\mathbf{1.00}$ & \\
\bottomrule
\end{tabular}
\end{table*}

\begin{table}[t]
\centering
\small
\setlength{\tabcolsep}{5pt}
\caption{\textbf{Zero-shot evaluation datasets.} All are real captured data and
disjoint from training. ``Boundary'' marks the datasets on which the
scale-invariant boundary F1 is additionally reported.}
\label{tab:eval_datasets}
\begin{tabular}{llrc}
\toprule
Dataset & Domain & \#Samples & Boundary \\
\midrule
DIODE~\cite{vasiljevic2019diode}   & indoor + outdoor          & $771$ & \\
KITTI~\cite{geiger2012kitti}   & outdoor (driving)         & $652$ & \\
NYUv2~\cite{silberman2012nyuv2}   & indoor                    & $654$ & \\
ETH3D~\cite{schops2017eth3d}   & indoor + outdoor          & $454$ & \\
HAMMER~\cite{jung2022hammer}  & indoor (multi-modal)      & $775$ & \checkmark \\
iBims-1~\cite{koch2018ibims} & indoor                    & $100$ & \checkmark \\
Booster~\cite{zamaramirez2022booster} & indoor (transparent/specular) & $38$  & \checkmark \\
\midrule
\textbf{Total} & & $\mathbf{3{,}444}$ & \\
\bottomrule
\end{tabular}
\end{table}

\begin{table}[h]
\centering
\small
\caption{\textbf{Training cost.} Number of epochs, H100 GPUs, and wall-clock time for the pre-training ($256\times256$) and fine-tuning ($512\times512$) stages of each variant.}
\label{tab:training}
\begin{tabular}{lccccccc}
\toprule
\multirow{2}{*}[-3.5pt]{Model} & \multirow{2}{*}[-3.5pt]{Param (M)} & \multicolumn{3}{c}{Pre-train ($256\times256$)} & \multicolumn{3}{c}{Fine-tune ($512\times512$)} \\
\cmidrule(lr){3-5} \cmidrule(lr){6-8}
 & & Epoch & GPUs & Time & Epoch & GPUs & Time \\
\midrule
PointDiT-B & 223  & 30 & 16 & 12h  & 8 & 64  & 2.5h \\
PointDiT-L & 771  & 30 & 16 & 21h  & 5 & 64  & 7h   \\
PointDiT-H & 1,807 & 30 & 64 & 22h  & 3 & 128 & 5.5h \\
\bottomrule
\end{tabular}
\end{table}

\section{More Evaluation Results}

\subsection{Per-Dataset Metrics}

\cref{tab:comparison} in the main paper reports metrics averaged over all seven evaluation datasets. For completeness, we provide the full per-dataset breakdown here: \cref{tab:per_dataset_point} for point map accuracy (Rel$^\text{p}$, $\delta_1^\text{p}$), \cref{tab:per_dataset_depth} for depth accuracy (Rel$^\text{d}$, $\delta_1^\text{d}$), and \cref{tab:per_dataset_bf1} for boundary sharpness (BF1). We split the point map and depth metrics into separate tables for readability, and report BF1 only on the three datasets whose ground truth supports boundary evaluation (HAMMER, iBims-1, Booster); per-dataset sample counts are given in \cref{tab:eval_datasets}. PointDiT variants use $4$ sampling steps, and the Avg column is the sample-weighted mean over all evaluation samples. Overall, PointDiT-H attains the best average depth accuracy, PointDiT achieves the sharpest boundaries on average, and PointDiT leads on the challenging HAMMER set across point map, depth, and boundary metrics. A notable exception is the datasets containing outdoor scenes: across KITTI, DIODE, and ETH3D, PointDiT is inferior to the strongest regression baselines such as MoGe and UniDepthV2. We attribute this gap mainly to the limited coverage of outdoor scenes in our synthetic training mixture, and we expect that introducing more outdoor datasets for training would further narrow it.

\begin{table}[t]
\begin{center}
\caption{\textbf{Per-dataset point map results.} Rel$^\text{p}$ $\downarrow$ and $\delta_1^\text{p}$ $\uparrow$ for each of the seven evaluation datasets at $512 \times 512$ resolution. PointDiT variants use $4$ sampling steps. The Avg column is the sample-weighted mean over all $3{,}444$ samples.}
\setlength{\tabcolsep}{3pt}
\resizebox{\textwidth}{!}{
\begin{tabular}{lcccccccccccccccc}
\toprule
\multirow{2}{*}{Method} & \multicolumn{2}{c}{DIODE} & \multicolumn{2}{c}{KITTI} & \multicolumn{2}{c}{NYUv2} & \multicolumn{2}{c}{ETH3D} & \multicolumn{2}{c}{HAMMER} & \multicolumn{2}{c}{iBims-1} & \multicolumn{2}{c}{Booster} & \multicolumn{2}{c}{Avg} \\
\cmidrule(lr){2-3} \cmidrule(lr){4-5} \cmidrule(lr){6-7} \cmidrule(lr){8-9} \cmidrule(lr){10-11} \cmidrule(lr){12-13} \cmidrule(lr){14-15} \cmidrule(lr){16-17}
 & Rel$^\text{p}$$\downarrow$ & $\delta_1^\text{p}$$\uparrow$ & Rel$^\text{p}$$\downarrow$ & $\delta_1^\text{p}$$\uparrow$ & Rel$^\text{p}$$\downarrow$ & $\delta_1^\text{p}$$\uparrow$ & Rel$^\text{p}$$\downarrow$ & $\delta_1^\text{p}$$\uparrow$ & Rel$^\text{p}$$\downarrow$ & $\delta_1^\text{p}$$\uparrow$ & Rel$^\text{p}$$\downarrow$ & $\delta_1^\text{p}$$\uparrow$ & Rel$^\text{p}$$\downarrow$ & $\delta_1^\text{p}$$\uparrow$ & Rel$^\text{p}$$\downarrow$ & $\delta_1^\text{p}$$\uparrow$ \\
\midrule
GeometryCrafter & 5.88 & 95.06 & 4.84 & 97.97 & 4.10 & 98.64 & 5.46 & 97.03 & 6.75 & 95.46 & 5.50 & 97.23 & 3.73 & 99.31 & 5.45 & 96.75 \\
PPD & 6.35 & 93.68 & 7.04 & 96.16 & 4.10 & 98.27 & 5.08 & 97.36 & 5.19 & 97.69 & 4.55 & 97.97 & 3.32 & 98.73 & 5.54 & 96.59 \\
Depth Pro & 6.36 & 94.11 & 7.33 & 95.97 & 4.08 & 98.48 & 5.90 & 96.85 & 5.27 & 98.03 & 4.54 & 98.11 & 2.87 & 99.81 & 5.71 & 96.71 \\
UniDepthV2 & 6.02 & 94.35 & \textbf{4.03} & \textbf{98.26} & 3.29 & 98.52 & 4.02 & 98.79 & 4.57 & 97.53 & 4.07 & 98.34 & 3.78 & 98.90 & 4.45 & 97.35 \\
DA3 & 5.47 & 93.68 & 6.37 & 96.72 & 3.67 & 98.20 & 3.44 & 98.20 & 4.60 & 96.95 & 4.08 & 97.81 & 2.92 & 99.03 & 4.77 & 96.63 \\
MoGe & \textbf{4.25} & \textbf{96.37} & 4.10 & 98.10 & 3.48 & 98.63 & \textbf{3.35} & \textbf{98.99} & 5.51 & 95.95 & 3.95 & 97.82 & 2.74 & 99.06 & \textbf{4.21} & 97.45 \\
MoGe-2 & 4.57 & 95.28 & 5.87 & 97.86 & \textbf{3.13} & \textbf{98.83} & 3.73 & 98.71 & 5.19 & 97.19 & 4.00 & 98.27 & \textbf{2.59} & \textbf{99.90} & 4.53 & 97.46 \\
\midrule
PointDiT-B & 6.62 & 93.83 & 5.65 & 97.08 & 5.05 & 97.54 & 6.67 & 96.97 & 5.67 & 98.52 & 4.87 & 97.69 & 3.81 & 99.58 & 5.85 & 96.80 \\
PointDiT-L & 5.69 & 95.05 & 5.18 & 97.55 & 4.35 & 98.08 & 4.95 & 97.96 & 4.26 & 99.18 & 4.25 & 98.14 & 3.15 & 99.88 & 4.85 & 97.55 \\
PointDiT-H & 5.12 & 95.97 & 5.09 & 97.86 & 3.95 & 98.40 & 4.56 & 98.30 & \textbf{3.52} & \textbf{99.54} & \textbf{3.91} & \textbf{98.57} & 3.01 & 99.87 & 4.40 & \textbf{98.02} \\
\bottomrule
\end{tabular}
}
\label{tab:per_dataset_point}
\end{center}
\end{table}

\begin{table}[t]
\begin{center}
\caption{\textbf{Per-dataset depth map results.} Rel$^\text{d}$ $\downarrow$ and $\delta_1^\text{d}$ $\uparrow$ for each of the seven evaluation datasets at $512 \times 512$ resolution. PointDiT variants use $4$ sampling steps. The Avg column is the sample-weighted mean over all $3{,}444$ samples.}
\setlength{\tabcolsep}{3pt}
\resizebox{\textwidth}{!}{
\begin{tabular}{lcccccccccccccccc}
\toprule
\multirow{2}{*}{Method} & \multicolumn{2}{c}{DIODE} & \multicolumn{2}{c}{KITTI} & \multicolumn{2}{c}{NYUv2} & \multicolumn{2}{c}{ETH3D} & \multicolumn{2}{c}{HAMMER} & \multicolumn{2}{c}{iBims-1} & \multicolumn{2}{c}{Booster} & \multicolumn{2}{c}{Avg} \\
\cmidrule(lr){2-3} \cmidrule(lr){4-5} \cmidrule(lr){6-7} \cmidrule(lr){8-9} \cmidrule(lr){10-11} \cmidrule(lr){12-13} \cmidrule(lr){14-15} \cmidrule(lr){16-17}
 & Rel$^\text{d}$$\downarrow$ & $\delta_1^\text{d}$$\uparrow$ & Rel$^\text{d}$$\downarrow$ & $\delta_1^\text{d}$$\uparrow$ & Rel$^\text{d}$$\downarrow$ & $\delta_1^\text{d}$$\uparrow$ & Rel$^\text{d}$$\downarrow$ & $\delta_1^\text{d}$$\uparrow$ & Rel$^\text{d}$$\downarrow$ & $\delta_1^\text{d}$$\uparrow$ & Rel$^\text{d}$$\downarrow$ & $\delta_1^\text{d}$$\uparrow$ & Rel$^\text{d}$$\downarrow$ & $\delta_1^\text{d}$$\uparrow$ & Rel$^\text{d}$$\downarrow$ & $\delta_1^\text{d}$$\uparrow$ \\
\midrule
GeometryCrafter & 3.74 & 96.89 & 3.73 & 98.20 & 2.99 & 98.82 & 3.26 & 98.38 & 3.86 & 97.24 & 2.95 & 98.06 & 2.25 & 99.14 & 3.52 & 97.84 \\
PPD & 4.69 & 95.86 & 4.81 & 97.47 & 3.57 & 98.38 & 4.15 & 97.69 & 2.57 & 99.33 & 3.11 & 98.44 & 2.56 & 99.21 & 3.88 & 97.78 \\
Depth Pro & 4.57 & 95.85 & 4.36 & 97.18 & 3.37 & 98.65 & 4.34 & 97.45 & 3.00 & 98.81 & 2.87 & 98.60 & 1.88 & \textbf{99.92} & 3.84 & 97.63 \\
UniDepthV2 & 3.62 & 97.11 & 3.25 & \textbf{98.44} & 2.60 & 98.84 & 2.65 & 99.00 & 2.23 & 99.42 & 2.46 & 98.49 & 2.27 & 98.98 & 2.86 & 98.52 \\
DA3 & 3.73 & 96.26 & 3.96 & 97.40 & 2.96 & 98.46 & 2.74 & 98.45 & 2.74 & 98.63 & 2.54 & 98.35 & 2.22 & 99.42 & 3.22 & 97.81 \\
MoGe & 3.17 & 97.32 & 3.26 & 98.33 & 2.64 & 98.90 & 2.53 & \textbf{99.17} & 3.75 & 96.92 & 2.57 & 98.14 & 2.21 & 99.16 & 3.10 & 98.01 \\
MoGe-2 & 2.99 & 97.61 & \textbf{3.16} & 98.36 & \textbf{2.58} & \textbf{98.95} & 2.71 & 99.03 & 3.13 & 98.52 & \textbf{2.26} & \textbf{98.83} & \textbf{1.77} & 99.75 & 2.90 & 98.45 \\
\midrule
PointDiT-B & 3.98 & 96.71 & 4.33 & 97.38 & 3.50 & 98.02 & 3.22 & 98.00 & 3.22 & 98.99 & 2.91 & 98.65 & 2.55 & 99.70 & 3.64 & 97.86 \\
PointDiT-L & 3.37 & 97.26 & 4.00 & 97.67 & 3.02 & 98.47 & 2.77 & 98.40 & 2.42 & 99.31 & 2.46 & 98.80 & 1.98 & 99.87 & 3.09 & 98.25 \\
PointDiT-H & \textbf{2.90} & \textbf{97.79} & 3.88 & 97.94 & 2.75 & 98.61 & \textbf{2.46} & 98.69 & \textbf{1.91} & \textbf{99.53} & 2.33 & 98.73 & 1.79 & 99.87 & \textbf{2.75} & \textbf{98.54} \\
\bottomrule
\end{tabular}
}
\label{tab:per_dataset_depth}
\end{center}
\end{table}

\begin{table}[t]
\begin{center}
\caption{\textbf{Per-dataset boundary sharpness (BF1 $\uparrow$).} BF1 is reported only on the three datasets whose ground truth supports boundary evaluation (HAMMER, iBims-1, Booster). PointDiT variants use $4$ sampling steps. The Avg column is the sample-weighted mean over these $913$ boundary-annotated samples.}
\small
\setlength{\tabcolsep}{10pt}
\begin{tabular}{lcccc}
\toprule
Method & HAMMER & iBims-1 & Booster & Avg \\
\midrule
GeometryCrafter & 3.34 & 11.25 & 13.78 & 4.64 \\
PPD & 7.60 & \textbf{16.95} & 22.78 & 9.26 \\
Depth Pro & 7.87 & 15.01 & 25.91 & 9.41 \\
UniDepthV2 & 6.22 & 9.54 & 14.91 & 6.94 \\
DA3 & 5.33 & 11.64 & 12.77 & 6.33 \\
MoGe & 4.28 & 11.87 & 16.39 & 5.61 \\
MoGe-2 & 5.97 & 14.36 & 18.28 & 7.40 \\
\midrule
PointDiT-B & 7.82 & 14.29 & 22.95 & 9.16 \\
PointDiT-L & \textbf{9.13} & 14.94 & 26.58 & \textbf{10.50} \\
PointDiT-H & 9.03 & 14.85 & \textbf{28.66} & 10.49 \\
\bottomrule
\end{tabular}
\label{tab:per_dataset_bf1}
\end{center}
\end{table}

\subsection{More Visualizations}

We show additional depth comparisons in \cref{fig:depth_compare}, and our PointDiT is significantly better in terms of reconstructing thin structures, transparent objects, and maintaining a more accurate relative scale across the global scene.

We provide more results in our project page: \url{https://haofeixu.github.io/pointdit}

\begin{figure*}[t!]
    \centering
    \vspace{-2mm}
    \includegraphics[width=\linewidth]{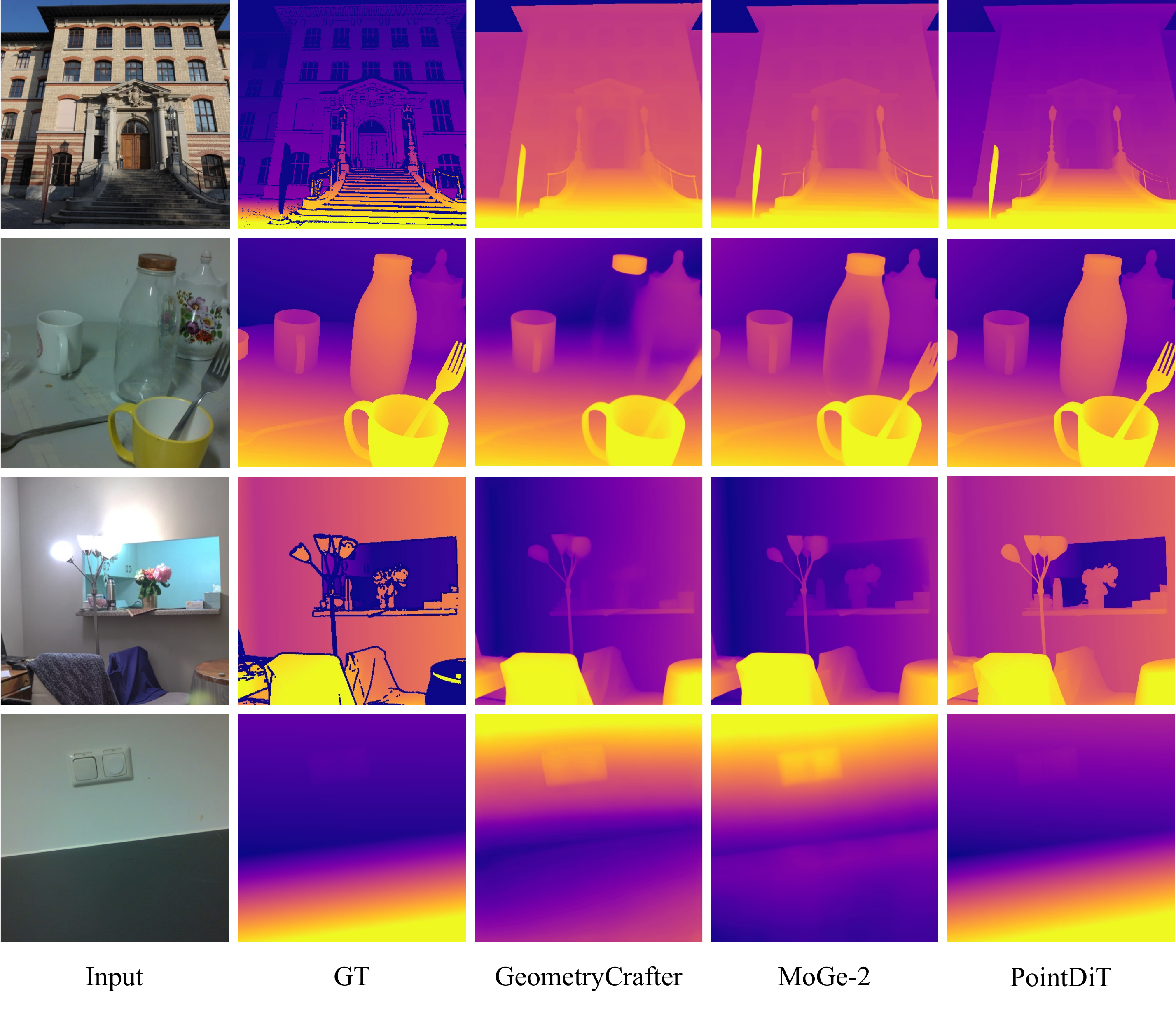}
    \caption{\textbf{Depth comparisons.} Our PointDiT is significantly better in terms of reconstructing thin structures (1st row), transparent objects (2nd rows), and maintaining a more accurate relative scale across the global scene (3rd and 4th rows). 
    The corresponding point map comparisons are provided in the main paper (\cref{fig:visual_compare}). 
    } 
    \label{fig:depth_compare}
\end{figure*}

\end{document}